\documentclass[10pt]{article} 
\usepackage{setspace}
\usepackage[preprint]{tmlr}

\usepackage{amsmath,amsfonts,bm}

\def\eqref#1{equation~\ref{#1}}

\def\1{\bm{1}}

\def\rd{{\textnormal{d}}}

\def\rw{{\textnormal{w}}}

\def\mD{{\bm{D}}}

\def\mG{{\bm{G}}}

\DeclareMathAlphabet{\mathsfit}{\encodingdefault}{\sfdefault}{m}{sl}
\SetMathAlphabet{\mathsfit}{bold}{\encodingdefault}{\sfdefault}{bx}{n}

\def\gD{{\mathcal{D}}}

\def\gH{{\mathcal{H}}}

\def\sH{{\mathbb{H}}}

\def\sR{{\mathbb{R}}}

\newcommand{\E}{\mathbb{E}}

\usepackage{amsmath,amsfonts}
\usepackage{amsthm}

\usepackage{array}
\usepackage[caption=false,font=normalsize,labelfont=sf,textfont=sf]{subfig}
\usepackage{textcomp}
\usepackage{stfloats}
\usepackage{verbatim}
\usepackage{graphicx}
\usepackage{cite}
\graphicspath{ {./} }
\usepackage[hyphens]{url}
\usepackage{hyperref}
\hypersetup{colorlinks=true,breaklinks=true,citecolor=blue!70, urlcolor=black}

\usepackage{tabularx,tabulary}
\usepackage{booktabs}
\usepackage{cleveref}
\usepackage{amssymb}
\usepackage{paralist}
\usepackage{etoolbox}
\usepackage[fleqn]{mathtools}
\usepackage{empheq}
\usepackage{tabularx,tabulary}
\usepackage{booktabs,siunitx}
\usepackage{multirow}%
\usepackage{dutchcal}

\usepackage{paralist}

\usepackage{bm}
\usepackage{subdepth}

\newcommand{\rsub}[1]{%
  \mathchoice
    {\text{\raisebox{0.1ex}{\scriptsize $(#1)$}}}
    {\text{\raisebox{0.1ex}{\scriptsize $(#1)$}}}
    {\text{\raisebox{0.1ex}{\tiny $(#1)$ }}}
    {\text{\raisebox{0.9ex}{\tiny $(#1)$}}}
}

\PassOptionsToPackage{usenames,dvipsnames}{xcolor}

\usepackage{afterpage}
\usepackage{placeins}

\DeclarePairedDelimiterX{\norm}[1]{\lVert}{\rVert}{\ifblank{#1}{\:\cdot\:}{#1}}
\DeclarePairedDelimiterX{\innerp}[2]{\langle}{\rangle}{#1,#2}
\DeclarePairedDelimiterXPP{\expectf}[1]{\mathbb{E}_{\substack{\sim\mathsf{0:T}}}}{[}{]}{}{\ifblank{#1}{\:\cdot\:}{#1}}
\DeclarePairedDelimiterXPP{\expect}[2]{\mathbb{E}_{#1}}{[}{]}{}{\ifblank{#2}{\:\cdot\:}{#2}}
\DeclarePairedDelimiterX{\abs}[1]{\lvert}{\rvert}{\ifblank{#1}{\:\cdot\:}{#1}}

\DeclarePairedDelimiterXPP{\expects}[2]{\displaystyle\mathop{{\mathbb{S}}}_{\scriptstyle{#1}}}{[}{]}{}{\ifblank{#2}{\:\cdot\:}{#2}}

\DeclarePairedDelimiterX\set[1]\lbrace\rbrace{#1}

\DeclarePairedDelimiterXPP{\expectt}[2]{{\mathbb{E}}_{\scriptstyle{#1}}}{\{}{\}}{}{\ifblank{#2}{\:\cdot\:}{#2}}

\DeclarePairedDelimiterXPP{\difft}[2]{{\mathbb{D}}_{\scriptstyle{#1}}}{\{}{\}}{}{\ifblank{#2}{\:\cdot\:}{#2}}

\title{A Trust-region Framework for Moment Estimation}

\author{\name Oluwasegun Somefun \email somefuno@oregonstate.edu
      }

\def\month{MM}  
\def\year{YYYY} 
\def\openreview{\url{https://openreview.net/forum?id=XXXX}} 

\usepackage{concmath}
\usepackage{newtxtext}

\begin{document}


\maketitle

\begin{abstract}
In this paper, we develop a trust-region framework for understanding the behavior of adaptive moment estimation mechanisms, such as \textsc{Adam}, in stochastic gradient optimization. Specifically, the magnitude of the update step associated with each individual parameter is constrained by a finite-order $p$-moment trust-region, with $p\ge1$. The resulting derivation leads to a family of learning-rate mechanisms based on second-moment estimation and normalized $p$-th-moment estimation. For $p=4$, this involves kurtosis estimation. Subsequent derivations provide a unified interpretation of moment-estimation-based normalization, learning-rate scheduling, momentum as a spectral first-order lowpass regularization, and operator-level spectral-norm normalization within a common trust-region framework. Preliminary experiments on GPT2-124M trained on FineWeb-Edu and TinyStories suggest that the fourth-moment realization provides its greatest benefit when trust-region constraints are weak. As progressively stronger trust-region controls are introduced, the second-moment realization becomes increasingly competitive, often achieving slightly lower validation loss than its corresponding fourth-moment realization.
\end{abstract}

\section{Introduction}

A natural question arising from the popular adaptive moment
estimation (\textsc{Adam}, \citet{kingmaAdamMethodStochastic2015})
mechanism in stochastic gradient optimization is whether its
moment-normalized update components can be understood from a rigorous trust-region
principle. 

In this paper, we address this question by developing a
moment-constrained trust-region control framework for stochastic gradient optimization. The central idea is to characterize update
magnitudes through explicit $p$-moment trust-region constraints. The resulting derivation leads to a family of learning-rate
mechanisms for $p \ge 1$, governed by a $p$-th moment trust-region constraint on the update step. In the special case $p=4$, the mechanism involves both moment estimation, and implicit estimation of the fourth root of the kurtosis associated with the gradient process. Hence, we refer loosely to realizations of this trust-region framework as \textsc{Gmake}. 

Beyond the derivation of a family of learning-rate mechanisms, the framework reveals that several mechanisms traditionally studied separately can be understood within a common trust-region perspective on the update step.~In particular, common learning-rate schedules arise naturally as solutions to
a trust-region variational problem, while momentum as a linear
iteration-invariant operator and
matrix-operator spectral-norm normalization can be viewed as complementary mechanisms for
progressively strengthening while preserving enforcement of an underlying
trust-region control on the update step process.
In this sense, the \textsc{Gmake} framework unifies moment-constrained trust-region optimization, learning-rate scheduling, moment estimation, momentum, and matrix-operator spectral-norm trust-region control within a
common framework. 

Here, our goal is not to establish the empirical superiority of a particular optimizer, but to systematically study the role of moment-based trust-region constraints in stochastic gradient methods. The principal contributions of this paper are:
\begin{asparaenum}[(i.)]
\item A $p$-moment-based trust-region formulation of the stochastic
gradient update step, for $p \ge 1$, leading to a generalized family of normalization mechanisms.

\item A unified interpretation of adaptive moment estimation,
learning-rate scheduling, spectral first-order lowpass filtering as momentum, and matrix-operator spectral-norm normalization as complementary trust-region mechanisms acting on different properties of the update step, while preserving
an underlying moment-based trust-region guarantees.
\end{asparaenum}

\subsection{Related Work}

The \textsc{Gmake} mechanism draws connections between several
research directions in stochastic gradient optimization, including
classic trust-region optimization, adaptive moment estimation, 
momentum, and matrix orthogonalization.

\paragraph{Adaptive moment estimation.}
Adaptive learning-rate algorithms such as \textsc{RMSProp} and
\textsc{Adam} normalize gradient components via second-moment estimates and has become the most popular optimizer in deep learning
\citep{kingmaAdamMethodStochastic2015,bottouOptimizationMethodsLargescale2018}.
In contrast, \textsc{Gmake} is derived from explicit
moment-constrained trust-region considerations. 
The resulting learning rule extends second-moment normalization
to a family of $p$-moment normalization mechanisms for $p\ge1$.

\paragraph{Trust-region, normalized-gradient methods, and learning-rate schedules.}
Trust-region methods regulate update magnitudes by constraining the
optimization step to lie within a prescribed neighborhood in which a
local model is considered reliable
\citep{parikhProximalAlgorithms2014, connTrustRegionMethods2000}.
Related ideas also appear in normalized-gradient methods, where update
directions are rescaled to control step magnitudes. Learning-rate
schedules \citep{bergsmaStraightZeroWhy2024, geRethinkingLearningRate2018} are also widely used to progressively reduce update
magnitudes throughout training, thereby influencing the effective size
of the optimization region explored by the algorithm. The formulation in
this paper differs in that the trust region is characterized through
moment constraints on the update itself, yielding a learning-rate
mechanism directly linked to the statistical properties of the
underlying gradient process. Furthermore, common learning-rate
schedules emerge naturally in this framework as solutions to a
trust-region variational control problem rather than as externally specified
heuristics.

\paragraph{Momentum and filtering.}
Momentum methods such as Heavy-Ball and Nesterov acceleration have
long been used to improve the stochastic gradient optimization process \citep{polyakAcceleratedGradientMethods2020,sutskeverImportanceInitializationMomentum2013, polyakConjugateGradientMethod1969}.
More recently, momentum mechanisms have
also been interpreted from a lowpass signal-processing perspective \citep{somefunFundamentalSignalProcessing2026, somefunAUTOSGMUnifiedLowpass2024b}, where they
introduce a smoothing effect on gradient sequences and attenuate
rapidly varying gradient components \citep{liPerformanceAnalysisMomentum2024a}.
These viewpoints provide useful intuition for understanding the noise-reduction properties of momentum-based optimization methods. In contrast, \textsc{Gmake} incorporates a passive linear, time-invariant operator that preferentially filters out high-frequency gradient fluctuations. Distinct from previous filtering interpretations, the operator is introduced as a mechanism for strengthening satisfaction of an existing moment-based trust-region constraint.

\paragraph{Matrix-aware and spectral-norm-constrained updates.}
Recent optimizer developments have emphasized the matrix structure of
neural-network parameters, leading to orthogonalized and
spectral-normalized update rules
\citep{bernsteinOldOptimizerNew2024}. Many such methods admit
interpretations in terms of alternative matrix norms or
operator-norm constraints \citep{largeScalableOptimizationModular2024}.
In contrast, spectral-norm constraints are introduced here as a secondary
mechanism for progressively tightening an existing moment-based
trust-region framework rather than as the primary update rule. The
matrix-operator form of \textsc{Gmake} is related to this literature in
that it applies spectral-norm normalization to a matrix-valued layer or parameter group composed of $p$-moment-normalized gradient components.

\subsection{A Trust-region problem}
A iteration $t$, consider a stochastic gradient algorithm applied to a parameter vector $\rw\rsub{t}\in\sR^{n'}$ of finite size $n'\ge 1$. For an arbitrary coordinate $i\in[1,n']$, let $w\rsub{t}\in\sR$ denote the associated parameter $\rw\rsub{t,i}$, and $g\rsub{t}\in\sR$ the associated gradient component obtained by minimizing a scalar-valued loss function $f\rsub{t}$ with respect to $w\rsub{t}$. Over a window of learning iterations $t\in[0,\tau]$, where $\tau \gg 1$, denote the update step as 
\begin{align}\label{sgmstep}
\Delta\rsub{t+1} = w\rsub{t+1} - w\rsub{t}.  
\end{align}
Throughout, we use $\E\{\cdot\}$ to denote expectation with respect to the underlying stochastic process generating the gradients.~No specific objective function, probabilistic structure, or distribution of the gradients is
assumed. We make only the following assumptions for all $t$:

\textbf{Assumption 1 (Lipschitz Regularity)}:~The function $f\rsub{t}$ is at least twice continuously differentiable, and both $f\rsub{t}$ and $g\rsub{t}$ are Lipschitz continuous in $w\rsub{t}$ \citep{bottouOptimizationMethodsLargescale2018}.

\textbf{Assumption 2 (Bounded $p$-th Moment)}:~The stochastic signal $g\rsub{t}$ satisfies $\|g\rsub{t}\|_\infty<\infty$.~Hence, for every finite $p\ge 1$,
$\E\{|g\rsub{t}|^p\}\le \|g\rsub{t}\|_\infty^p <\infty$.
Further, we assume that each iteration of the update step (\ref{sgmstep}) is constructed to at least, ensure $\E\{\abs{\Delta\rsub{t+1}}\} \le \mu$. In other words, the expected magnitude of the update step is bounded within a maximum trust region radius $\mu > 0$. The trust-region radius of the update step is characterized by its $p$-th moment
\begin{align}\label{trrule}
   \E\{\abs{\Delta\rsub{t+1}}^p\} = \delta_p^p\rsub{t} \le \mu^p,
\end{align}
where $\mu$ is also the maximum allowable update step-size, while ${\delta_p}\rsub{t}$ is a trust-region radius shaping function governing variation of the update step, with ${\delta_p}\rsub{t} \le \mu$, with ${\delta_p}\rsub{t}\to 0$ as $t \to \tau$. 

\subsection{Overview of the Trust-region Framework}
The basic form of the learning algorithm, with a variational shaping function $\varsigma\rsub{t} \in [0, 1]$, and $\varsigma\rsub{t}\to 0$ as $t \to \tau$ is
\begin{align}
  &\delta_p\rsub{t} = \mu\,\varsigma\rsub{t}^{\frac{1}{p}},\label{alg1-start}\\
  &\tilde{g}\rsub{t} = \frac{g\rsub{t}}{\E\{|g\rsub{t}|^2\}^\frac{1}{2}},\\
  &\bar{g}\rsub{t} = \frac{\tilde{g}\rsub{t}}{\E\{|\tilde{g}\rsub{t}|^p\}^\frac{1}{p}},\\    
  &\Delta\rsub{t+1} = -\delta_p\rsub{t}\, \bar{g}\rsub{t}.\label{alg1-end}
\end{align}
Next, let $\sH\colon\!\ell^2 \!\to\! \ell^2$ be a linear, iteration-invariant operator, applied on $g\rsub{t}$, with spectral-norm $\Vert \sH \Vert_\infty \le 1$. The spectral regularized learning algorithm can be expressed as
\begin{align}
  &\delta_p\rsub{t} = \mu\,\varsigma\rsub{t}^{\frac{1}{p}},\label{alg2-start}\\
  &v\rsub{t} = \sH\{g\rsub{t}\}, \label{alg2-lpf}\\
  &\tilde{g}\rsub{t} = \frac{v\rsub{t}}{\E\{|g\rsub{t}|^2\}^\frac{1}{2}},\\
  &\bar{g}\rsub{t} = \frac{\tilde{g}\rsub{t}}{\E\{|\tilde{g}\rsub{t}|^p\}^\frac{1}{p}},\\    
  &\Delta\rsub{t+1} = -\delta_p\rsub{t}\, \bar{g}\rsub{t}\label{alg2-end}.
\end{align}
Further, let
$\bar{\mG}\rsub{t}$ denote a full-rank $n \times m$ matrix composed of normalized gradient components $\bar{g}\rsub{t}$ associated
with a single layer or parameter group $l$, where $1 < n \le m$. Denote the update step in matrix form as
$\mD\rsub{t+1}$. Now, suppose that, in order to strengthen direct enforcement of the individual trust-region constraints $\E\{\abs{\Delta\rsub{t+1}}\} \le \mu$,
we additionally introduce a bound on the layer's spectral norm as
$\norm{\mD\rsub{t+1}}_2 \le \mu$.
The algorithm then takes the form
\begin{align}
  &\delta_p\rsub{t} = \mu\,\varsigma\rsub{t}^{\frac{1}{p}},\label{alg3-start}\\
  &v\rsub{t} = \sH\{g\rsub{t}\},\\
  &\tilde{g}\rsub{t} = \frac{v\rsub{t}}{\E\{|g\rsub{t}|^2\}^\frac{1}{2}},\\
  &\bar{g}\rsub{t} = \frac{\tilde{g}\rsub{t}}{\E\{|\tilde{g}\rsub{t}|^p\}^\frac{1}{p}},\\    
  &\tilde{\mG}\rsub{t}
      =
      \bigl(
      \bar{\mG}\rsub{t}\bar{\mG}\rsub{t}^{\intercal}
      \bigr)^{-\frac{1}{2}}
      \bar{\mG}\rsub{t},\label{alg3-max}\\
  &\mD\rsub{t+1} = -\delta_p\rsub{t}\,\tilde{\mG}\rsub{t}\label{alg3-end}.
\end{align}

\subsection{Organization of the Paper}

The remainder of this paper is organized as follows.~Section~\ref{sec-gmake}  derives the basic form of \textsc{Gmake}, establishes its
moment-constrained properties, and connection to learning-rate schedules.~Section~\ref{sec-spectral-lp} introduces a
trust-region-preserving spectral lowpass regularization operator. Then, Section~\ref{sec-spectral-mat} develops a matrix-operator form that additionally enforces spectral-norm trust-region constraint.~Section~\ref{sec-pract} notes practical implementation strategies for the required statistical estimators, and Section~\ref{sec-numexp} presents numerical
experiments and discussions comparing the \textsc{Gmake} forms. Current limitations and gaps are discussed in Section~\ref{sec-lim}. Finally, in Section~\ref{sec:concl}, this paper is concluded.

\section{{\textsc{Gmake}}}\label{sec-gmake}
In its basic form, the stochastic gradient update in (\ref{sgmstep}) can be defined as
\begin{align}
  \Delta\rsub{t+1} = w\rsub{t+1} - w\rsub{t} = -\alpha\rsub{t}\,g\rsub{t},
\end{align}
where $\alpha\rsub{t}$ is a non-negative real-valued learning rate function of iteration $t$. Our goal is to design a learning-rate mechanism $\alpha\rsub{t}$ that ensures the $\Delta\rsub{t+1}$ satisfies the trust-region problem involving its $p$-th moment
$\E\{\abs{\Delta\rsub{t+1}}^p\} \le \mu^p$, so that
$\E\{\abs{\Delta\rsub{t+1}}\} \le \mu$.

\subsection{Learning-rate mechanism}
Recall that $\delta_p^p\rsub{t} = \E\{\abs{\Delta\rsub{t+1}}^p\}$. Substituting the update step into the trust-region constraint yields
\begin{align}\label{step1}
  \E\{\abs{\Delta\rsub{t+1}}^p\} = \E\{\abs{-\alpha\rsub{t}\,g\rsub{t}}^p\} = \alpha^p\rsub{t}\,\E\{\abs{g\rsub{t}}^p\} = \delta_p^p\rsub{t}.
\end{align}
Solving for $\alpha\rsub{t}$ gives the learning-rate mechanism
\begin{align}\label{lr1}
  \alpha\rsub{t} = \frac{\delta_p\rsub{t}}{\E\{\abs{g\rsub{t}}^p\}^{\frac{1}{p}}}.
\end{align}
Define $\tilde{g}\rsub{t} = g\rsub{t}/\E\{\abs{g\rsub{t}}^2\}^{\frac{1}{2}}$ as the root-mean-square (rms) normalized form of the input signal $g\rsub{t}$,
then it follows that ${\E\{\abs{g\rsub{t}}^p\}^{\frac{1}{p}}} = \E\{\abs{\tilde{g}\rsub{t}}^p\}^{\frac{1}{p}} \,\E\{\abs{g\rsub{t}}^2\}^{\frac{1}{2}} $. The learning rate (\ref{lr1}) can be equivalently expressed as
\begin{align}\label{lr2}
  \alpha\rsub{t} = \frac{\delta_p\rsub{t}}{\E\{\abs{\tilde{g}\rsub{t}}^p\}^{\frac{1}{p}}\,\E\{\abs{g\rsub{t}}^2\}^{\frac{1}{2}} },
\end{align}
and the update step (\ref{step1}) becomes
\begin{align}\label{step2}
  \Delta\rsub{t+1} = -\delta_p\rsub{t}\,\frac{\tilde{g}\rsub{t}}{\E\{\abs{\tilde{g}\rsub{t}}^p\}^{\frac{1}{p}}}.
\end{align}
In particular, for $p=4$, the learning rate function has a meaningful interpretation of being composed of terms related to the second-moment and the fourth-root of the kurtosis of $g\rsub{t}$.
Next, denote $\bar{g}\rsub{t} = {\tilde{g}\rsub{t}}/{\E\{\abs{\tilde{g}\rsub{t}}^p\}^{\frac{1}{p}}}$, then the update step (\ref{lr2}) can be expressed as $\Delta\rsub{t+1} = -\delta_p\rsub{t}\,\bar{g}\rsub{t}$, which enforces $\E\{\abs{\bar{g}\rsub{t}}^p\}=1$, and because $\delta_p^p\rsub{t} \le \mu^p$, yields $\E\{\abs{\Delta\rsub{t+1}}^p\} \le \mu^p$. 
As a result, for $p \ge 1$, Lyapunov's inequality implies
\begin{align}\label{emaxss-ineq}
\E\{\abs{\Delta\rsub{t+1}}\}
\le
\E\{\abs{\Delta\rsub{t+1}}^p\}^{\frac 1 p}
\le \mu.
\end{align}
The update step $\Delta\rsub{t+1} = -\delta_p\rsub{t}\,\bar{g}\rsub{t}$ then directly satisfies both trust-region constraints on the $p$-th moment and expected maximum magnitude. The resulting update algorithm sequentially follows the outlined equations (\ref{alg1-start})--(\ref{alg1-end}). Further, let $\rd\rsub{t+1}\in\sR^{n'}$ denote the corresponding
update step vector for which $\Delta\rsub{t+1}$ is an $i$-th component. Since the $p$-moment normalization acts
independently on each parameter, the trust-region constraint
(\ref{emaxss-ineq}) implies that, 
\begin{align}
\E\|\rd\rsub{t+1}\|_r^r
&=
\sum_{i=1}^{n'}
\E\!\left\{
\left|
\rd\rsub{t+1,i}
\right|^r
\right\}
\le
\sum_{i=1}^{n'} \mu^r
=
n'\mu^r
\end{align}
for any $1 \le r \le p$. Since $0<\mu<1$ and $r\ge1$, it also follows that
$\mu^r \le \mu$, yielding a looser upper bound
$\E\|\rd\rsub{t+1}\|_r^r\le n'\mu$.
Therefore, the $p$-th moment-based trust-region constraint directly controls
the expected magnitude of the vectorized update step in every
$\ell_r$ norm, where $1\le r\le p$, and in particular implies
$\E\|\rd\rsub{t+1}\|_p^p \le n'\mu^p$.

For convenience, denote $\kappa_p\rsub{t}=\E\{\abs{\tilde{g}\rsub{t}}^p\}^{\frac{1}{p}}$, $\lambda_2\rsub{t}=\E\{\abs{g\rsub{t}}^2\}^{\frac{1}{2}}$, and note that $\kappa_p\rsub{t} \ge 1$, we can re-write the underlying learning-rate mechanism (\ref{lr2}) as
\begin{align}\label{lr2-simple}
  \alpha\rsub{t} = \frac{\delta_p\rsub{t}}{\kappa_p\rsub{t}\, \lambda_2\rsub{t}} \le \frac{\mu}{\lambda_2\rsub{t}},
\end{align}
The overall algorithm in (\ref{alg1-start}) -- (\ref{alg1-end}) can be interpreted as using $\lambda_2\rsub{t}$ to first normalize the variance scale of $g\rsub{t}$, and then applying $\kappa_p\rsub{t}$ to yield the fully normalized $\bar{g}\rsub{t}$, whose $p$-th moment satisfies the trust-region constraint.
Ideally, the special case, $p=2$ yields $\kappa_2\rsub{t} = 1$, and (\ref{lr2-simple}) reduces to the same mechanism
used in \textsc{RMSProp} and \textsc{Adam}, where $g\rsub{t}$ is
normalized solely by $\lambda_2\rsub{t}$. 
However, during estimation, when $\kappa_2\rsub{t} \ne 1$, the algorithm applies an additional
normalization by $\kappa_2\rsub{t}$ to compensate for scaling errors
that remain after the $\lambda_2\rsub{t}$ normalization. As a result,
this approach can be interpreted as a principled refinement of the
normalization mechanism employed by both \textsc{RMSProp} and \textsc{Adam}.

Also, for $p\ne2$, when the normalization factor
$\kappa_p\rsub{t}$ remains close to its minimum value
of one, the resulting learning-rate mechanism becomes
increasingly similar to the $p=2$ case. 
Conversely, when higher-order gradient fluctuations
is more pronounced and $\kappa_p\rsub{t}$ departs further
from one, the additional normalization acts as a stronger
trust-region control mechanism.

\subsection{Learning-rate Schedules as Trust-region shaping functions}
The trust-region problem requires that $\delta_p^p\rsub{t}\le \mu^p$ for all $t$, and that $\delta_p^p\rsub{t} \to 0$ as $t\to \tau$. This can be interpreted as progressively tightening the trust-region constraints over the learning window to enforce asymptotic stability of the update step. 
In order to obtain the smoothest possible variation of the trust-region radius across iterations over the learning horizon $t\in[0,\tau]$, we can define a total variational energy functional minimization problem directly on $\delta_p^p\rsub{t}$, 
\begin{align}\label{tvefcn1}
  \min_{0 \le \delta_p^p\rsub{t} \le \mu^p}\:\gD(\delta_p,\tau) = \sum_{t=1}^{\tau} \bigl[\delta_p^p\rsub{t} - \delta_p^p\rsub{t-1} \bigr]^2,
\end{align}
subject to fixed boundary values: the terminal boundary value $\delta_p^p\rsub{\tau}=0$, together with at least one specified initial boundary value satisfying $\delta_p^p\rsub{0}\le \mu^p$. 
The energy minimization problem (\ref{tvefcn1}) represents the canonical first-order smoothness functional that penalize rapid variations in $\delta_p^p\rsub{t}$, the trust-region radius between successive iterations. Equivalently, (\ref{tvefcn1}) encourages $\delta_p^p\rsub{t}$ to remain as close as possible to $\delta_p^p\rsub{t-1}$ throughout learning.
Consequently, $\delta_p\rsub{t}$ varies no faster than necessary while remaining consistent with the asymptotic stability requirement, thereby producing the smoothest trust-region variation consistent with the boundary values.

Furthermore, (\ref{tvefcn1}) is strictly convex in $\delta_p^p\rsub{t}$, and therefore has a unique minimizer. Differentiating with respect to any interior index $i\in\{1,\ldots,\tau-1\}$, and setting ${d\gD}/{d\delta_p^p\rsub{i}}=0$ results in the second-order homogeneous linear difference equation with constant coefficients
\begin{align}\label{order2}
  \delta_p^p\rsub{i+1} -2\,\delta_p^p\rsub{i} + \delta_p^p\rsub{i-1} = 0.
\end{align}
Let $\Delta\,\delta_p^p\rsub{i+1} = \delta_p^p\rsub{i+1} - \delta_p^p\rsub{i}$, then (\ref{order2}) corresponds to the second-order finite difference operator $\Delta\bigl(\Delta\,\delta_p^p\rsub{i+1}\bigr) = \Delta^2\,\delta_p^p\rsub{i+1} = 0$. The corresponding characteristic equation is $z^2 -2\,z + 1=(z-1)^2 = 0$ which has $1$ as a root with multiplicity 2. Therefore, the general solution to (\ref{order2}) must be a polynomial in $t$, of at most degree $1$ \citep{elaydiIntroductionDifferenceEquations2005, luekerTechniquesSolvingRecurrences1980}, defined as
\begin{align}\label{gsol-order2}
    \delta_p^p\rsub{t} = a_0 + a_1\,t.
\end{align}
The complete solution can then be obtained by plugging in at least $b \ge 2$ fixed boundary values: the fixed terminal value, and at least one fixed initial value, leading to $b-1$ sub-interval(s) of the learning window.

\subsection{2-point boundary value}
Consider the initial and terminal boundary values: $\delta_p^p\rsub{0}= \mu^p$, $\delta_p^p\rsub{\tau}=0$ over the single sub-interval $0\le t \le \tau$
Solving (\ref{gsol-order2}), subject to $\delta_p^p\rsub{0}= \mu^p$, $\delta_p^p\rsub{\tau}=0$ gives $a_0=\mu^p$, and $a_1=-\mu^p/\tau$. Therefore, the complete solution to (\ref{order2}) subject to the 2-point boundary value is
\begin{align}\label{csol-2pt}
    \delta_p^p\rsub{t} = \mu^p\,\Bigl(1-\frac{t}{\tau}\Bigr), \quad 0\le t \le \tau,
\end{align}
which is known as the linear decay schedule. We can simplify this to a normalized functional form. Let $u=t/\tau$, define an input function $x(u) \coloneq u \in [0,1]$, then 
\begin{align}\label{ncsol-2pt}
    \delta_p^p\rsub{t} = \mu^p\,\bigl(1-x(u)\bigr), \quad 0\le u \le 1.
\end{align}

\subsection{3-point boundary value}
Consider the three fixed boundary values: $\delta_p^p\rsub{0}= 0$, $\delta_p^p\rsub{k}=\mu^p$, $\delta_p^p\rsub{\tau}=0$, where $0\le k < \tau$. This corresponds to two sub-intervals. On the first interval over $0\le t \le k$, the solution is obtained by solving (\ref{gsol-order2}) subject to $\delta_p^p\rsub{0}= 0$, $\delta_p^p\rsub{k}=\mu^p$. On the second interval over $k\le t\le\tau$, the solution is obtained by solving (\ref{gsol-order2}) subject to $\delta_p^p\rsub{k}=\mu^p$, $\delta_p^p\rsub{\tau}=0$. 
Combining the piecewise solutions, the complete solution to (\ref{order2}) subject to the 3-point boundary value is
\begin{align}\label{csol-3pt}
    \delta_p^p\rsub{t} = \begin{cases}
      \mu^p\,\frac{t}{k}, & 0 \le t \le k,\quad k > 0\\
      \mu^p\,\frac{\tau-t}{\tau-k} & k \le t \le \tau,\quad k\ge 0.
    \end{cases}
  \end{align}
Simplifying further, define $u=t/\tau \in [0,1]$, a normalized rise-time $m=k/\tau \in [0,1)$, and the reparameterized input function 
\begin{align}\label{csol-3pt-x}
    x(u;m) = \begin{cases}
      u, & m=0,\\
      \max\bigl\{\frac{m-u}{m},\, \frac{u-m}{1-m}\bigr\}, & 0 < m < 1,
    \end{cases}
\end{align}
then equivalently, (\ref{csol-3pt}) is
\begin{align}\label{ncsol-3pt}
    \delta_p^p\rsub{t} = \mu^p\,\bigl(1-x(u;m)\bigr), \quad 0\le u \le 1, \quad 0\le m < 1.
\end{align}
For its first sub-interval, the hat-shaped (or triangular-like) mapping (\ref{csol-3pt}) corresponds to a warmup schedule that reaches $\mu^p$ at time $m$, and then a decaying schedule to zero in the second sub-interval. 
Importantly, observe that both (\ref{ncsol-2pt}) and (\ref{ncsol-3pt}) share the same underlying functional form. When $m=0$, it follows that $x(u;m)=x(u;0)=u$, therefore (\ref{ncsol-2pt}) is a special case of (\ref{ncsol-3pt}) when $m=0$. 

\subsection{4-point boundary value} 

Consider the four fixed boundary values: $\delta_p^p\rsub{0}= 0$, $\delta_p^p\rsub{k}=\mu^p$, $\delta_p^p\rsub{k'}=\mu^p$, $\delta_p^p\rsub{\tau}=0$, where $0\le k \le k' < \tau$. This corresponds to three sub-intervals. On the first interval over $0\le t \le k$, the solution is obtained by solving (\ref{gsol-order2}) subject to $\delta_p^p\rsub{0}= 0$, $\delta_p^p\rsub{k}=\mu^p$. On the second interval over $k\le t\le k'$, the solution is obtained by solving (\ref{gsol-order2}) subject to $\delta_p^p\rsub{k}=\mu^p$, $\delta_p^p\rsub{k'}=\mu^p$. Finally, for the third interval over $k'\le t\le\tau$, the solution is obtained by solving (\ref{gsol-order2}) subject to $\delta_p^p\rsub{k'}=\mu^p$, $\delta_p^p\rsub{\tau}=0$. 
Combining the three piecewise solutions, the complete solution to (\ref{order2}) subject to the 4-point boundary value, when simplified is
\begin{align}\label{csol-4pt}
    \delta_p^p\rsub{t} = \begin{cases}
      \mu^p\,\frac{t}{k}, & 0 \le t \le k, \quad k>0\\
      \mu^p\, & k \le t \le k', \quad k,\,k'\ge 0,\\
      \mu^p\,\frac{\tau-t}{\tau-k'} & k' \le t \le \tau, \quad k'\ge 0.
    \end{cases}
\end{align}
Define $u=t/\tau \in [0,1]$, and let $m=k/\tau \in [0,1)$ indicate the rise time and subsequently, the start of a plateau of width $\varepsilon = (k'-k)/\tau \in [0,1)$, so that $k=m\tau$, and $k'=(m+\varepsilon)\tau$. Then
the reparameterized input function $x(r;m,\varepsilon) \in [0,1]$ for (\ref{csol-4pt}) is
\begin{align}
x(u;m)\coloneq
\begin{cases}
\max\bigl\{0, \tfrac{u-\varepsilon}{1-\varepsilon}\bigr\}, & m=0,\\
\max\bigl\{\tfrac{m-u}{m},\,0,\,
\tfrac{u-(m+\varepsilon)}{1-(m+\varepsilon)}\bigr\},
& 0<m<1,
\end{cases}
\end{align} 
and equivalently (\ref{csol-4pt}) can be expressed as
\begin{align}\label{ncsol-4pt}
    \delta_p^p\rsub{t} = \mu^p\,\bigl(1-x(u;m,\varepsilon)\bigr), \quad 0\le u \le 1, \quad 0\le m < 1.
\end{align}
The resulting function (\ref{csol-4pt}) then corresponds in the first sub-interval to a warmup schedule that rises to $\mu^p$, and then a plateau in the second sub-interval, followed by a decaying schedule to zero in the third sub-interval. This leads a trapezoidal shape, and exactly a warmup-stable-decay schedule. 

For $m=0,\varepsilon=0$, it follows that $x(u;m,\varepsilon)=x(u;0,0)=u$, therefore (\ref{ncsol-2pt}) is a special case of (\ref{ncsol-4pt}). 
Similarly, when $m\ne0,\varepsilon=0$, it follows that $x(u;m,\varepsilon)=x(u;m,0)=x(u;m)$, therefore (\ref{ncsol-3pt}) is a special case of (\ref{ncsol-4pt}). 
Importantly, both (\ref{ncsol-2pt}), (\ref{ncsol-3pt}) and (\ref{ncsol-4pt}) share the same underlying functional form, with (\ref{ncsol-4pt}) having a more general input reparameterization function.

Finally, denote $\delta_p^p\rsub{t} = \mu^p\,\varsigma\rsub{t}$, where $\varsigma\rsub{t} = 1-x(u;m,\varepsilon)$, then
\begin{align}
  \delta_p\rsub{t} = \mu\,\varsigma^{\frac{1}{p}}\rsub{t}
\end{align}
is the unique minimizer of the first-order total variational energy functional, which satisfies $\delta_p\rsub{t}\le \mu$, $\delta_p\rsub{t} \to 0$ as $t\to \tau$ subject to $2 \le b \le 4$ fixed boundary points.
\begin{figure}[h]
  \centering
  \includegraphics[width=0.33\textwidth]{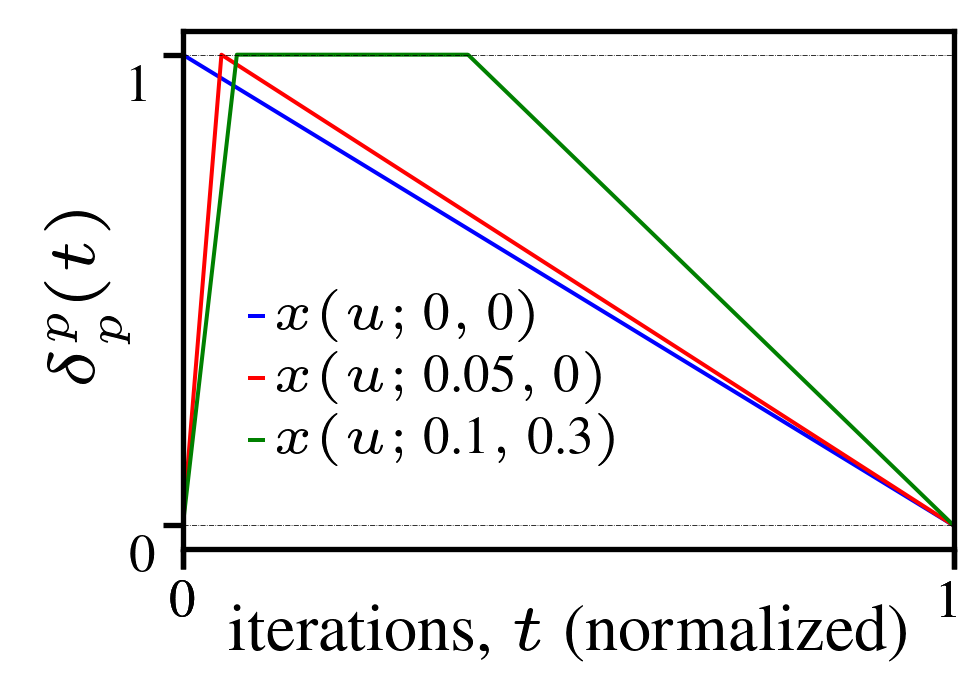}\hfil
  \includegraphics[width=0.33\textwidth]{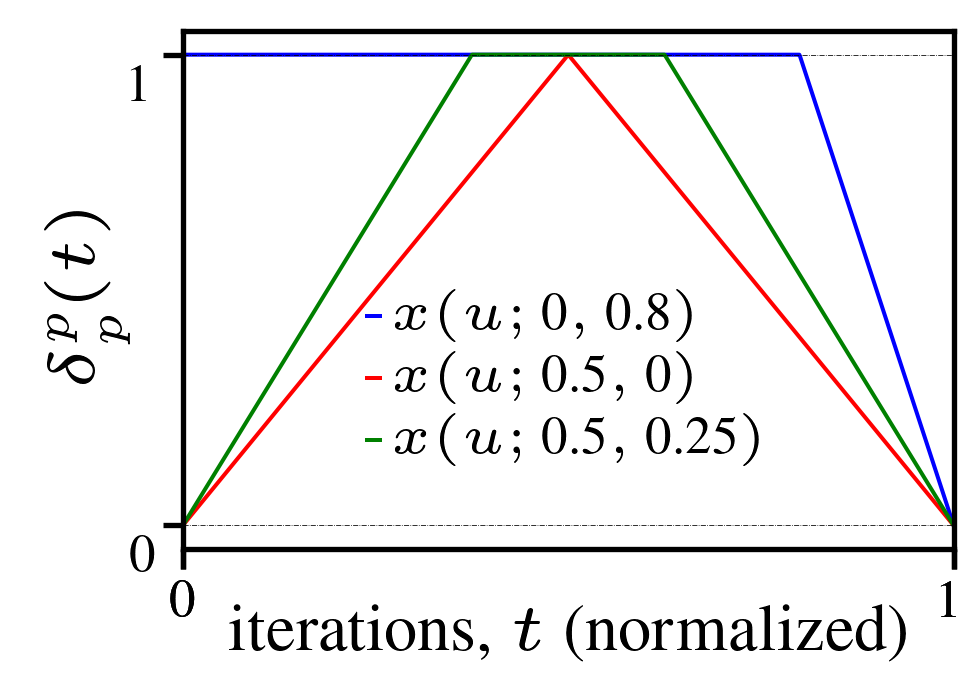}
  \vspace{-2ex}
  \caption{Different reparameterizations of the boundary-value problem on $\delta_p^p\rsub{t}$, with $\mu=1$ lead to the shapes of common learning-rate schedules, sharing the same linear mapping $1-x(u;m,\varepsilon)$.}
  \label{fig:bvfig}
\end{figure}

\textbf{Remarks:} 
The preceding analysis shows that many commonly used learning-rate schedules arise naturally as solutions to a trust-region variational problem. By treating $\delta_p^p\rsub{t}$ as a varying trust-region radius and minimizing its total variation subject to fixed boundary values, we obtain schedules that evolve as smoothly as possible, while still satisfying the asymptotic requirement $\delta\rsub{t}\to0$ as $t\to\tau$. 

The resulting $p$-th root linear decay, warmup-decay, and warmup-stable-decay schedules are illustrated in Figure~{\ref{fig:bvfig}}. They correspond respectively to 2-point, 3-point, and 4-point boundary-value problems, yet all share a common variational structure.
Under this interpretation, schedule design is no longer an independent heuristic, but rather a principled consequence of solving a first-order variational optimization problem subject to trust-region boundary constraints. More generally, alternative variational functionals would produce different classes of schedules \citep{somefunFundamentalSignalProcessing2026}, each reflecting the structure of the underlying variational principle. Since $\delta_{p}\rsub{t}$ directly controls the allowable update magnitude, learning-rate scheduling serves as a complementary mechanism through which the trust-region constraint of the update step is gradually tightened over the learning horizon via a variational shaping function.

\subsection{Underlying Taylor-series Model}\label{sec:taylseries}
The role of the $p$-moment trust-region constraint can also be viewed from the perspective of a Taylor-series local approximation model.
Let the update step (\ref{step1}) generated by (\ref{lr2}) satisfy the
$p$-moment trust-region constraint
\begin{align}\label{pmomtr}
\E\{\abs{\Delta\rsub{t+1}}^p\}
=
\delta_p^p\rsub{t}
\le \mu^p,
\end{align}
where $\delta_p\rsub{t}\to 0$ as $t\to\tau$.
For $p>r>0$, Lyapunov's inequality implies
\begin{align}\label{lowpcntrl-ineq}
\E\{\abs{\Delta\rsub{t+1}}^r\}^{\frac1r}
\le
\E\{\abs{\Delta\rsub{t+1}}^p\}^{\frac1p}
=
\delta_p\rsub{t}.
\end{align}
Consequently, all lower moments up to order $p$ are simultaneously controlled,
\begin{align}\label{allmomctl}
\E\{\abs{\Delta\rsub{t+1}}^r\}
\le
\delta_p^r\rsub{t},
\qquad r\le p.
\end{align}
In particular, since
$\Delta\rsub{t+1}=-\tfrac{\delta_p\rsub{t}}{\kappa_p\rsub{t}}\,\bar{g}\rsub{t}$, $\E\{\abs{\bar{g}\rsub{t}}^2\}=1$ and $\kappa_p\rsub{t}\ge1$, it follows that
\begin{align}\label{2momctl}
\E\{\abs{\Delta\rsub{t+1}}^2\}
=
\tfrac{\delta_p^2\rsub{t}}{\kappa_p^2\rsub{t}}
\le
\delta_p^2\rsub{t}.
\end{align}
Therefore, a $p$-moment trust-region constraint would control every moment
entering a Taylor-series expansion up to degree $p$. Consider
the update $w\rsub{t+1}=w\rsub{t}+\Delta\rsub{t+1}$.
Let $g\rsub{t}$ and $h\rsub{t}$ be the Lipschitz continuous first and second derivatives of $f\rsub{t}$ with respect to $w\rsub{t}$.
By Taylor's theorem, the second-order Taylor model associated with the expected loss decrease can be expressed as
\begin{align}\label{taylor2mdl}
\E\{\Delta f\rsub{t+1}\}
=
\E\{g\rsub{t}\Delta\rsub{t+1}\}
+
\frac12\,\E\{h\rsub{t}\Delta^2\rsub{t+1}\}
+
O\!\left(
\E\{\abs{\Delta\rsub{t+1}}^3\}
\right),
\end{align}
where
$\Delta f\rsub{t+1}=f\rsub{t+1}-f\rsub{t}$, and (\ref{taylor2mdl}) follows directly from Taylor's remainder theorem.
Thus, there exists a constant $\mathbcal{c}>0$ such that the third-order remainder
satisfies
$|R\rsub{3}| \le \mathbcal{c}|\Delta\rsub{t+1}|^3$.
Consequently,
$|E\{R\rsub{3}\}| \le E\{|R\rsub{3}|\}
\le \mathbcal{c}\,E\{|\Delta\rsub{t+1}|^3\}$,
and therefore $E\{R\rsub{3}\}=O(E\{|\Delta\rsub{t+1}|^3\})$.

Denote $m\rsub{t} = \E\{h\rsub{t}\bar{g}\rsub{t}^2\}$. Applying the moment bounds on $\Delta\rsub{t+1}$, by using (\ref{allmomctl}) and (\ref{2momctl}) in (\ref{taylor2mdl}) yields
\begin{align}
\E\{\Delta f\rsub{t+1}\}
\le
-\tfrac{\delta_p\rsub{t}}{\kappa_p\rsub{t}}\,
\lambda_2\rsub{t}
+
\tfrac{1}{2}\,
\tfrac{\delta_p^2\rsub{t}}{\kappa_p^2\rsub{t}}\,m\rsub{t}
+
O(\delta_p^3\rsub{t}).
\end{align}
Observe that the second-order Taylor model contains
$\Delta\rsub{t+1}$, $\Delta^2\rsub{t+1}$, and a leading neglected remainder of order
$\Delta^3\rsub{t+1}$. To simultaneously control the first-order term,
second-order term, and cubic remainder, it is sufficient that $p\ge3$.
The choice $p=4$ is also attractive because it is the smallest
moment order that simultaneously controls the variance and tail heaviness behavior of the update process.
Although larger values $p>4$ provide
progressively stronger higher-order moment control, they do not fundamentally
improve the underlying second-order Taylor model. Moreover, the benefits of controlling moments beyond order four often diminish faster
than the difficulty of reliably estimating them in practice \citep{ramachandranCrossfluctuationPhaseTransitions2025,jondeauOptimalPortfolioAllocation2006}. 

More importantly, the distinction between a fixed and a vanishing
trust-region constraint can also be noted. If $\delta_p\rsub{t}=\mu$, we have
$\E\{\abs{\Delta\rsub{t+1}}^p\}\le\mu^p$, then all
lower moments remain bounded, and the Taylor approximation error remains
controlled but generally non-vanishing. In contrast, the vanishing
trust-region schedule $\delta_p\rsub{t}\to0$ implies
\begin{align}
\E\{\abs{\Delta\rsub{t+1}}^r\}
\to 0,
\qquad r\le p,
\end{align}
causing every lower-order moment of the update step to vanish.~As a
result,
$
{O(\delta_p^2\rsub{t})}/{O(\delta_p\rsub{t})}\to0$,
$
{O(\delta_p^3\rsub{t})}/{O(\delta_p\rsub{t})}\to0$,
so that both the second-order correction and higher-order Taylor
remainder become asymptotically negligible relative to the leading
first-order term. Consequently,
\begin{align}
\E\{\Delta f\rsub{t+1}\}
\le
-\tfrac{\delta_p\rsub{t}}{\kappa_p\rsub{t}}\,
\lambda_2\rsub{t}
+
O(\delta_p^2\rsub{t}),
\end{align}
and the local Taylor model becomes progressively more accurate as
learning proceeds.

\textbf{Remarks:} Therefore, the role of the vanishing $p$-moment trust-region constraint
extends beyond merely controlling update magnitudes. By driving all lower
moments of the update step to vanish, it simultaneously contracts the
trust region and enforces asymptotic validity of the local Taylor-series
model used to describe the expected decrease in the loss.
As learning proceeds and $\delta_p\rsub{t}\to0$, the expected
loss decrease becomes increasingly well described by the first-order
term, while higher-order terms become asymptotically negligible.

\section{Trust-Region-Preserving Spectral Regularization}\label{sec-spectral-lp}
Denote $v\rsub{t} = \sH\{g\rsub{t}\}$, so that the stochastic gradient update in (\ref{sgmstep}) becomes
\begin{align}\label{sgmstepv}
  \Delta\rsub{t+1} = w\rsub{t+1} - w\rsub{t} = -\alpha\rsub{t}\,v\rsub{t},
\end{align}
\begin{align}\label{lr1v}
  \alpha\rsub{t} = \frac{\delta_p\rsub{t}}{\E\{\abs{v\rsub{t}}^p\}^{\frac{1}{p}}}.
\end{align}
Let $\sH\colon\!\ell^2 \!\to\! \ell^2$ be a linear iteration-invariant first-order filter characterized by a pole location $\beta\in\sR$ and zero location $\gamma\in\sR$, with transfer function
\begin{align}\label{eq:folpf}
\sH(z)
=
\eta
\frac{1-\gamma z^{-1}}
     {1-\beta z^{-1}},
\qquad
\eta=\frac{1-\beta}{1-\gamma},
\end{align}
where $z=e^{j\omega}$ over frequencies $\omega\in[-\pi,\pi]$, and $\sH(1)=1$.  The squared magnitude, frequency response  is
\begin{align}\label{pgain}
\abs{\sH(e^{j\omega})}^2=
\frac{(1-\beta)^2}{(1-\gamma)^2}
\frac{1+\gamma^2-2\gamma\cos\omega}
     {1+\beta^2-2\beta\cos\omega}.
\end{align}
To show that the system's worst-case gain, its $\gH_\infty$ norm \citep{boydLinearControllerDesign1991b} defined as $\norm{\sH}_\infty = \max_{\omega}\,\norm{\sH(e^{j\omega})}_2 = \max_{\omega}\,\abs{\sH(e^{j\omega})} \le 1,\,\forall\, \omega\in[-\pi,\pi]$, it suffices to show conditions for which the denominator of (\ref{pgain}) is always greater than or equal to its numerator.
Rearranging terms, this implies
\begin{align}
\Delta(\omega)
&=
(1-\gamma)^2(1+\beta^2-2\beta\cos\omega)
-
(1-\beta)^2(1+\gamma^2-2\gamma\cos\omega) \ge 0\nonumber \\
&=
2(\beta-\gamma)(1-\beta\gamma)(1-\cos\omega) \ge 0.
\label{eq:delta}
\end{align}
Note that in (\ref{eq:delta}), the following holds: $1-\cos\omega\ge 0$ is non-negative for all $\omega$; factor $\beta-\gamma\ge 0$ if $\abs{\gamma} < \abs{\beta}$; and factor $1-\beta\gamma>0$ if $|\beta|<1$ and $|\gamma|<1$.
Therefore, a sufficient condition for the system's $\norm{\sH}_\infty \le 1$ is
\begin{align}\label{pamramrng}
0 \le \beta < 1,\qquad \abs{\gamma} < \abs{\beta}.
\end{align}
This design condition ensures strong attenuation of high-frequency components in $g\rsub{t}$, and ensures that the linear operator is non-expansive in its induced $\ell_2$ norm. By definition of the induced norm \citep{boydLinearControllerDesign1991b}, this implies $\Vert v\rsub{t} \Vert_2 \le \Vert g\rsub{t} \Vert_2$, then, $v\rsub{t}$ denotes an output sequence, whose total finite energy can be no larger than that of its input $g\rsub{t}$. 

Also, by the convolution property of the linear time-invariant operator, $v\rsub{t}=\sum_{k=0}^{t}\mathbcal{h}\rsub{k}\,g\rsub{t-k}$ holds, where $\mathbcal{h}\rsub{k}$ is the impulse response of $\sH$ \citep{oppenheimDiscreteTimeSignalProcessing2010a} defined as $\mathbcal{h}\rsub{0}=\eta$, $\mathbcal{h}\rsub{k}=\eta(\beta-\gamma)\beta^{k-1}$, and due to (\ref{pamramrng}) $\mathbcal{h}\rsub{k}\ge0$ for all $k\ge 0$ is satisfied. It follows that $\norm{\mathbcal{h}}_{1} =1$, and that $\norm{v\rsub{t}}_{\infty} \le \norm{g\rsub{t}}_{\infty}$. Therefore, by definition, the linear-time invariant operator $\sH$ is also non-expansive in its induced $\ell_\infty$ norm.  

Using Assumption 2, if
$\E\{\abs{\Delta\rsub{t+1}}^p\} = \alpha^p\rsub{t}\,\E\{\abs{g\rsub{t}}^p\}
\le \alpha^p\rsub{t}\,\norm{g\rsub{t}}^p_{\infty} \le \delta_p^p\rsub{t}$ for the unfiltered update step, then
after the spectral regularization, 
\begin{align}
\E\{\abs{\Delta\rsub{t+1}}^p\} = \alpha^p\rsub{t}\,\E\{\abs{v\rsub{t}}^p\} \le \alpha^p\rsub{t}\,\norm{v\rsub{t}}^p_{\infty} \le \alpha^p\rsub{t}\,\norm{g\rsub{t}}^p_{\infty}  \le \delta_p^p\rsub{t}.  
\end{align}
Consequently,
all trust-region guarantees regarding the allowable update magnitude, inherited from the underlying $p$-moment mechanism are preserved and may become strictly tighter after regularization.~Therefore, $\sH$ does not introduce an additional trust-region constraint. Rather, it improves the quality of the gradient signal used to generate the update step by smoothing high-frequency gradient fluctuations before the learning-rate mechanism is applied.
We will refer to this effect providing strong attenuation of high-frequency spectral components in $g\rsub{t}$, while having a spectral norm less than one at all frequencies $\omega\in[-\pi,\pi]$, as a trust-region-preserving spectral lowpass regularization of $g\rsub{t}$.

\subsection{Variance reduction factor}
A useful measure of the output variance is the system $\gH_2$ norm, its average energy over all frequencies $\omega\in[-\pi,\pi]$,
\begin{align}\label{eq:H2factor}
\norm{\sH}_2^2=
\frac{1}{2\pi}\int_{-\pi}^\pi\,\norm{\sH(e^{j\omega})}_F^2\,d\omega =
\frac{1}{2\pi}\int_{-\pi}^\pi\,\abs{\sH(e^{j\omega})}^2\,d\omega
=\frac{1-\beta}{1+\beta}\frac{1+\gamma^2-2\beta\gamma}{(1-\gamma)^2}
\end{align}
In particular, for a white noise input signal, this quantifies the residual noise variance after filtering \citep{orfanidisIntroductionSignalProcessing1995, boydLinearControllerDesign1991b}. Therefore, minimizing (\ref{eq:H2factor}) can be interpreted as maximizing variance reduction. Given $0 \le \beta < 1$, (\ref{eq:H2factor}) can be used to place $\gamma$. Differentiating with respect to $\gamma$, $\frac{d}{d\gamma}\|\sH\|_2^2$ is strictly positive, and increasing over the admissible range $-\beta<\gamma<\beta$. Consequently, the  minimum variance placement for the zero location $\gamma$ is
\begin{align}
\boxed{
\gamma_{\rm MVR}=-\beta}.
\end{align}
Substituting $\gamma=-\beta$ into $\eta=\frac{1-\beta}{1-\gamma}$, yields
$\eta_{\rm MVR}=
\frac{1-\beta}{1+\beta}$. As $\beta\to1$,
$\eta_{\rm MVR}=o(1-\beta)$. Note that if $\eta=1-\beta$, then $\gamma=0$.
Either way, maximum variance reduction is obtained at the cost of an increasingly small $\eta$.
The normalization gain $\eta$ governs how 
the filter, realized in (\ref{fcanrealss}) responds to changes in its input signal. In this case, the input signal $g\rsub{t}$ carries information about the local optimization landscape.
However as $\eta\to 0$, maximal variance reduction, smoothing effect is increased, leading to less responsiveness to meaningful changes in $g\rsub{t}$. The reverse is the case as $\eta\to 1$.
Consequently, in this setting, maximizing variance reduction alone is not an efficient filter design in terms of simultaneously reducing variance while preserving ability to track changes in its input signal.


\subsection{Variance reduction per unit gain}

A more efficient design tradeoff is to balance the variance measure relative to its responsiveness measure, defined as
\begin{align}\label{eq:vrg-criterion}
J(\gamma)
=\frac{\norm{\sH}_2^2}{\eta}=
\frac{1+\gamma^2-2\beta\gamma}
     {(1+\beta)(1-\gamma)}.
\end{align}
Minimizing (\ref{eq:vrg-criterion}) is equivalent to maximizing the filter's variance reduction per unit gain. Differentiating with respect to $\gamma$ yields $(1-\gamma)^2=
2(1-\beta)$,
whose feasible root is
$\gamma=
1-\sqrt{2(1-\beta)}$.
Subject to $-\beta\le\gamma\le\beta$, the optimal solution becomes
\begin{align}\label{eq:gamma_star}
\boxed{
\gamma_{\rm VRG}
=
\max
\left\{
-\beta,
\;
1-\sqrt{2(1-\beta)}
\right\}.
}
\end{align}
The transition occurs when
$-\beta=
1-\sqrt{2(1-\beta)}$,
which yields a threshold
$\beta=\sqrt5-2\approx0.236$ related to the golden ratio \citep{pronzatoEstimationSpectralBounds2013}.
Therefore, for small $0\le\beta<\sqrt5-2$, the variance-per-unit-gain optimum $\gamma_{\rm VRG}=-\beta$ coincides with the maximum variance reduction choice. Otherwise, as $\beta\to 1$, for $\sqrt5-2\le\beta<1$, the variance-per-unit-gain optimum increasingly favors preservation of responsiveness,
$\gamma_{\rm VRG} = 1-\sqrt{2(1-\beta)}$.

\subsection{Canonical Filter Realization}
A direct difference equation realization of the first-order filter (\ref{eq:folpf}) is
\begin{align}\label{vdf}
  v\rsub{t} = \beta \, v\rsub{t-1} + \eta\,\big(g\rsub{t}  - \gamma \, g\rsub{t-1}  \big).
\end{align}
In general, (\ref{vdf}) admits the controllable canonical realization
\citep{oppenheimDiscreteTimeSignalProcessing2010a, smithPoleZeroAnalysis2007},
\begin{align}\label{fcanreal}
\begin{aligned}
q\rsub{t} &= \beta\,q\rsub{t-1}+g\rsub{t},\\
v\rsub{t} &= \eta\bigl(q\rsub{t}-\gamma q\rsub{t-1}\bigr),
\end{aligned}
\end{align}
where $q\rsub{t}$ is the state generated during realization of the filter. Additional refinements,
such as dividing the filter output by $1-\beta^t$, compensate for
the transient-response bias of the state during the initial
iterations.

To make the lowpass regularization of $g\rsub{t}$ explicit,
define $\tilde q\rsub{t}=(1-\beta)\,q\rsub{t}$. Since
$\eta=\tfrac{1-\beta}{1-\gamma}$ and
$1-\eta=\tfrac{\beta-\gamma}{1-\gamma}$, it follows that
$\eta(\beta-\gamma)q\rsub{t-1}
=(1-\eta)\tilde q\rsub{t-1}$.
Substituting into (\ref{fcanreal}) yields
\begin{align}\label{fcanrealss}
\begin{aligned}
\tilde q\rsub{t}
&= \beta\,\tilde q\rsub{t-1}+(1-\beta)\,g\rsub{t},
\\
v\rsub{t}
&= \eta\,g\rsub{t}
 +(1-\eta)\,\tilde q\rsub{t-1},
\end{aligned}
\qquad\Leftrightarrow\qquad
\begin{aligned}
e\rsub{t}
&= \beta\,e\rsub{t-1}
 + \bigl(g\rsub{t-1}-g\rsub{t}\bigr),
\\
v\rsub{t}
&= g\rsub{t}
 +(1-\eta)\,e\rsub{t},
\end{aligned}
\end{align}
where the right-hand-side representation follows from the change
of variables $e\rsub{t}=\tilde q\rsub{t-1}-g\rsub{t}$.

\subsection{Beyond Variance Reduction}
From (\ref{fcanrealss}), recall $e\rsub{t} = \beta\,e\rsub{t-1} + (g\rsub{t-1}-g\rsub{t})$. Using the unit-delay operator $z^{-1}$, and defining the first-order finite-difference operator $\Delta = 1-z^{-1}$, we have that $(g\rsub{t}-g\rsub{t-1}) = (1-z^{-1})g\rsub{t} = \Delta g\rsub{t}$. The $e\rsub{t}$ state recursion with zero initialization $e\rsub{0}=0$, and repeated substitutions can then be expressed as
\begin{align}
e\rsub{t}
=
-
\sum_{k=0}^{t-1}
\beta^k
z^{-k}(1-z^{-1})g\rsub{t} =
-\sum_{k=0}^{t-1}
\beta^k z^{-k}\Delta g\rsub{t},
\label{eq:error_series}
\end{align}
and hence $v\rsub{t} = g\rsub{t} + (1-\eta)\,e\rsub{t}$ is
\begin{align}\label{eq:v_series}
v\rsub{t}
=
g\rsub{t}
-
(1-\eta)
\sum_{k=0}^{t-1}
\beta^k z^{-k}\Delta g\rsub{t}.
\end{align}
Equation~(\ref{eq:error_series}) is an exponentially weighted sum of the delayed first-order finite differences of $g\rsub{t}$. 
Each term $z^{-k}(1-z^{-1})g\rsub{t}$,
is generated by delayed applications of the first-order finite-difference
operator $(1-z^{-1})$ on $g\rsub{t}$, where $z^{-k}g\rsub{t}=g\rsub{t-k}$, and $z^{-k}(1-z^{-1})g\rsub{t} = (1-z^{-1})g\rsub{t-k}$.
Hence, (\ref{eq:v_series}) can be interpreted as the gradient component $g\rsub{t}$ being regularized through a correction term formed from an exponentially weighted sum of its delayed first-order finite differences.

Consequently, recent ($0 \le k \ll t-1$) rapidly varying terms $z^{-k}\Delta g\rsub{t}$ produce larger correction terms, whereas recent slowly varying or constant terms produce smaller corrections.
Therefore, beyond typical noise variance reduction, the lowpass filter can be viewed as performing a \emph{regularization} of the gradient sequence by selectively penalizing large iteration-to-iteration changes in each gradient component $g\rsub{t}$.

\subsection{Heavy-ball and Nesterov momentum}
\begin{figure}[h]
  \centering
  \includegraphics[width=0.33\textwidth]{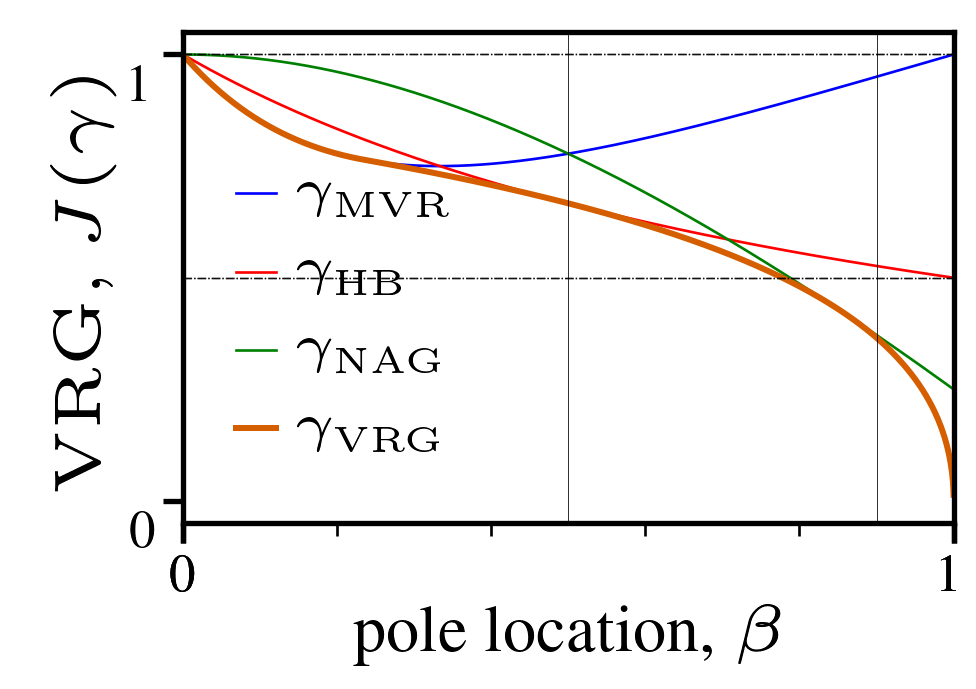}
  \vspace{-2ex}
  \caption{Variance reduction per unit gain. For $\beta > \sqrt{5}-2$, $\gamma_{\rm MVR}$ clearly becomes inefficient. Observe that $\gamma_{\rm NAG}$ closely matches $\gamma_{\rm VRG}$ as $\beta \gg 0.5$, especially $\beta \approx 0.8-0.9$, whereas $\gamma_{\rm HB}$ closely matches $\gamma_{\rm VRG}$ for $\beta \approx 0.4-0.6$. At the cross-over point $\beta \approx 0.71$, $\gamma_{\rm NAG}$ overtakes $\gamma_{\rm HB}$ as an efficient design. }
  \label{fig:pz}
\end{figure}

Although, the design choice (\ref{eq:gamma_star}), makes the linear operator to efficiently balance variance reduction with responsiveness, the same filter admits other closely related configurations. More generally, given $0\le\beta<1$, by appropriate choices of $\gamma$, the linear operator can be used to recover both Heavy-ball and Nesterov momentum. Algebraically, analyzing (\ref{fcanreal}), Heavy-ball momentum can be recovered by selecting $\gamma_{\rm HB}=0$,
\begin{align}\label{hb}
  \begin{aligned}
  &v\rsub{t}  = \beta\,v\rsub{t-1}  + (1-\beta)\,g\rsub{t},
  \end{aligned}
\end{align}
while Nesterov momentum corresponds to
$\gamma_{\rm NAG}=\frac{\beta}{1+\beta}$,
\begin{align}\label{nestform}
  \begin{aligned}
  &q\rsub{t}  = \beta \, q\rsub{t-1}  +  (1-\beta)\,g\rsub{t} \\
  &v\rsub{t}  = \beta\,q\rsub{t}  + (1-\beta)\,g\rsub{t}.
  \end{aligned}
\end{align}
Given $0\le\beta<1$, four special operating points of the first-order lowpass filter are therefore
\begin{align}
\boxed{
\gamma_{\rm MVR}=-\beta,
\qquad
\gamma_{\rm HB}=0,
\qquad
\gamma_{\rm NAG}=\frac{\beta}{1+\beta},
\qquad
\gamma_{\rm VRG}=\max\{-\beta, 1-\sqrt{2(1-\beta)}\}.
}
\end{align}
More specifically, within the same trust-region-preserving first-order filter family, both Heavy-ball and Nesterov momentum can be viewed as specific operating choices of $\gamma$. Recall that this family satisfies the spectral norm less than one design condition (\ref{pamramrng}). Generally, lowpass filter design involves making tradeoffs among several competing effects \citep{orfanidisIntroductionSignalProcessing1995}. Maximum variance reduction causes a simultaneous maximum reduction in the filter's normalization gain. While this improves noise suppression, it also effectively reduces the maximum step-size of the update process, which can potentially lead to slower adaptation and less responsiveness to the underlying optimization dynamics.~As shown in Figure~\ref{fig:pz}, Nesterov momentum achieves more variance reduction per unit gain than Heavy-ball for $\beta \gtrapprox 0.71$. 
Overall, this unified spectral regularization interpretation of classic momentum methods through the pole-zero locations of a first-order lowpass filter, also corresponds to design tradeoffs between maximal variance reduction and responsiveness of the filtering dynamics to the local optimization landscape. Outlined in (\ref{alg2-start})--(\ref{alg2-end}) is the trust-region-preserving regularized algorithm, with (\ref{alg2-lpf}) being the only addition to the basic algorithm.

\section{Matrix-operator form}\label{sec-spectral-mat}

In Section~\ref{sec-gmake}, a vectorized parameter group was considered. In contrast, now let $n'=nm$, and initially denote
$\tilde{\mG}\rsub{t}:=\bar{\mG}\rsub{t}$ as a full-rank
$n\times m$ matrix composed of normalized gradient components
$\bar g\rsub{t}$ associated with a single layer or parameter
group $l$, where $1<n\le m$. Then in matrix form, the update
step of (\ref{alg2-start})--(\ref{alg2-end}) can be denoted as
\begin{align}
\mD\rsub{t+1}
=
-\delta_p\rsub{t}\,\tilde \mG\rsub{t}.
\end{align}
The $p$-moment normalization acts on each individual $ij$ entry
of the matrix, where
$1\le i\le n$ and $1\le j\le m$, and therefore enforces
$
\E\{|\tilde \mG_{ij}\rsub{t}|^r\}\le 1
$ for $1\le r\le p$. It follows that $\E\{|{\mD_{ij}}\rsub{t+1}|^r\}\le\mu^r$ is equivalent to $\E\{|\Delta\rsub{t+1}|^r\}\le\mu^r$.
Viewing the update as a matrix, the corresponding operator gain
satisfies,
\begin{align}\label{sporig}
\E\|\mD\rsub{t+1}\|_2^2
\le
\E\|\mD\rsub{t+1}\|_F^2
=
\delta_p^2\rsub{t}
\sum_{ij}^{mn}
\E\{|\tilde \mG_{ij}\rsub{t}|^2\}
\le
\mu^2 mn.
\end{align}
Therefore, although the entrywise $p$-moment trust-region
controls the magnitude of the vectorized update, it cannot
directly control the operator gain of the corresponding matrix
update group. In particular, the bound (\ref{sporig}) exceeds $\mu^2$.
An algebraic consequence is that if
$
\E\|\mD\rsub{t+1}\|_2
\le
\mu,
$
then
$
\E|{\mD_{ij}}\rsub{t+1}|
\le
\E\|\mD\rsub{t+1}\|_2
\le
\mu.
$
Therefore, a tighter trust-region control can be obtained
by constraining the entire matrix update group at the spectral
level according to
\begin{align}
\|\mD\rsub{t+1}\|_2
\le
\mu',
\end{align}
where $\mu\le\mu'<\mu\sqrt{mn}$. In this range, the special case
$\mu'=\mu$ corresponds to the strongest trust-region constraint
relative to (\ref{sporig}).
An effective way to control the spectrum of a matrix is through an orthogonalization step \citep{highamFunctionsMatricesTheory2008}, such as,
\begin{align}
\tilde{\mG}\rsub{t}
=
\left(
\bar{\mG}\rsub{t}\bar{\mG}\rsub{t}^\intercal
\right)^{-\frac{1}{2}}
\bar{\mG}\rsub{t}.
\end{align}
By the singular-value decomposition
$
\bar{\mG}\rsub{t}
=
U\rsub{t}\Sigma\rsub{t}V\rsub{t}^\intercal,
$
it follows that
$
\tilde{\mG}\rsub{t}
= U\rsub{t}V\rsub{t}^\intercal,
$
and therefore
$
\|\tilde{\mG}\rsub{t}\|_2=1.
$
The layer or matrix group update step
$
\mD\rsub{t+1}
=
-\delta_p\rsub{t}\,\tilde{\mG}\rsub{t}
$
then satisfies
\begin{align}
\|\mD\rsub{t+1}\|_2
=
\delta_p\rsub{t}\|\tilde{\mG}\rsub{t}\|_2
=
\delta_p\rsub{t}
\le
\mu.
\end{align}
Since $\tilde{\mG}\rsub{t}=U\rsub{t}V\rsub{t}^\intercal$, each $ij$ entry, directly satisfies
$
|\tilde{\mG}_{ij}\rsub{t}|
=
|u_{i}^\intercal \rsub{t} v_{j}\rsub{t}|
\le
\|u_{i}\rsub{t}\|_2
\|v_{j}\rsub{t}\|_2
\le 1,
$
and therefore
$|\mD_{ij}\rsub{t+1}|\le\mu$, equivalent to $|\Delta\rsub{t+1}|\le\mu$ is also directly satisfied.

This matrix-operator form of the regularized algorithm is outlined in (\ref{alg3-start})--(\ref{alg3-end}), with (\ref{alg3-max}) introduced. 
Moment normalization does not directly control the operator gain (largest singular value) of the matrix group, while spectral normalization does not regulate the statistical variability of its individual entries. Therefore, the preceding $p$-moment normalization and the subsequent spectral normalization
are complementary trust-region mechanisms that operate on different properties of the update matrix group. 
By directly controlling the operator gain of the entire matrix update group to be less than the maximum trust-region constant $\mu$, spectral normalization strengthens the underlying trust-region framework. 

\section{Practical realizations}\label{sec-pract}

This section discusses practical realizations of the statistical
expectations appearing in \textsc{Gmake}, together with a practical
realization of the principal matrix inverse square-root required by
the matrix-operator form.

\textbf{Matrix inverse square-root}.
For computing the principal matrix inverse square-root in
(\ref{alg3-max}), we adopt the efficient polynomial recursion of
\citet{lakicComputationMatrixKth1998}.

\textbf{Linear statistical estimators}.
\textsc{Gmake} requires online estimates that ensure
$\lambda_2\rsub{t} > 0$ and $\kappa_p\rsub{t} \ge 1$.
Since these expectations are unavailable \emph{a priori},
they must be replaced by recursive estimators.  Here, we adopt classical estimators from the stochastic approximation literature whose statistical properties are already well established \citep{zoubirRobustStatisticsSignal2018a, lehmannTheoryPointEstimation2005, jamesEstimationQuadraticLoss1961}.
A common practical estimator is the exponentially weighted moving
average (EMA)
\begin{align}
m\rsub{t}
=
\rho\,m\rsub{t-1}
+
(1-\rho)\,x\rsub{t},
\qquad
0<\rho<1,
\end{align}
which may be used to estimate both quantities by selecting
$x\rsub{t}=|g\rsub{t}|^2$, $m\rsub{0}=0$, or
$x\rsub{t}=|\bar g\rsub{t}|^p$, $m\rsub{0}=1$,
respectively.
For the normalized $p$-th moment
$\kappa_p\rsub{t}$, sometimes a possible structured estimate can be unity. In such situations, a linear shrinkage estimator (LSE) such as
\begin{align}
m\rsub{t}
=
\rho\,m_0
+
(1-\rho)\,x\rsub{t},
\qquad
0<\rho<1,
\end{align}
with prior value $m_0=1$ may also provide an alternative realization
\citep{jamesEstimationQuadraticLoss1961}.~When the prior is
accurate, the LSE can reduce estimator variance
\citep{ledoitWellconditionedEstimatorLargedimensional2004} while
avoiding the additional memory state required by a second EMA.
Both estimators are also subject to initial transient bias from the true underlying mean value \citep{box2015time,goodwinAdaptiveFilteringPrediction1984}. The typical
bias correction may be applied by dividing the estimates by
$1-\rho^t$. For both, values of $\rho$ very close to unity typically
provide robust averaging
\citep{haykinAdaptiveFilterTheory2014,ljung1983theory}.

\section{Numerical Experiments}\label{sec-numexp}
The algorithms defined by
(\ref{alg1-start})--(\ref{alg1-end}),
(\ref{alg2-start})--(\ref{alg2-end}), and
(\ref{alg3-start})--(\ref{alg3-end})
will be referred to as the basic, spectrally filtered, and
matrix-operator forms of \textsc{Gmake}, respectively. These
correspond to progressively stronger realizations of the
underlying trust-region framework. As proof of concept, we
compare the three forms for both $p=2$ and $p=4$. All experiments, were repeated three times, and the average training loss and validation loss curves are reported. The GPT2-124M model processes 8192 tokens per iteration. Several observations can be made. 

\subsection{FineWeb-Edu}
\begin{figure}[h]
  \centering
  \includegraphics[width=0.48\textwidth]{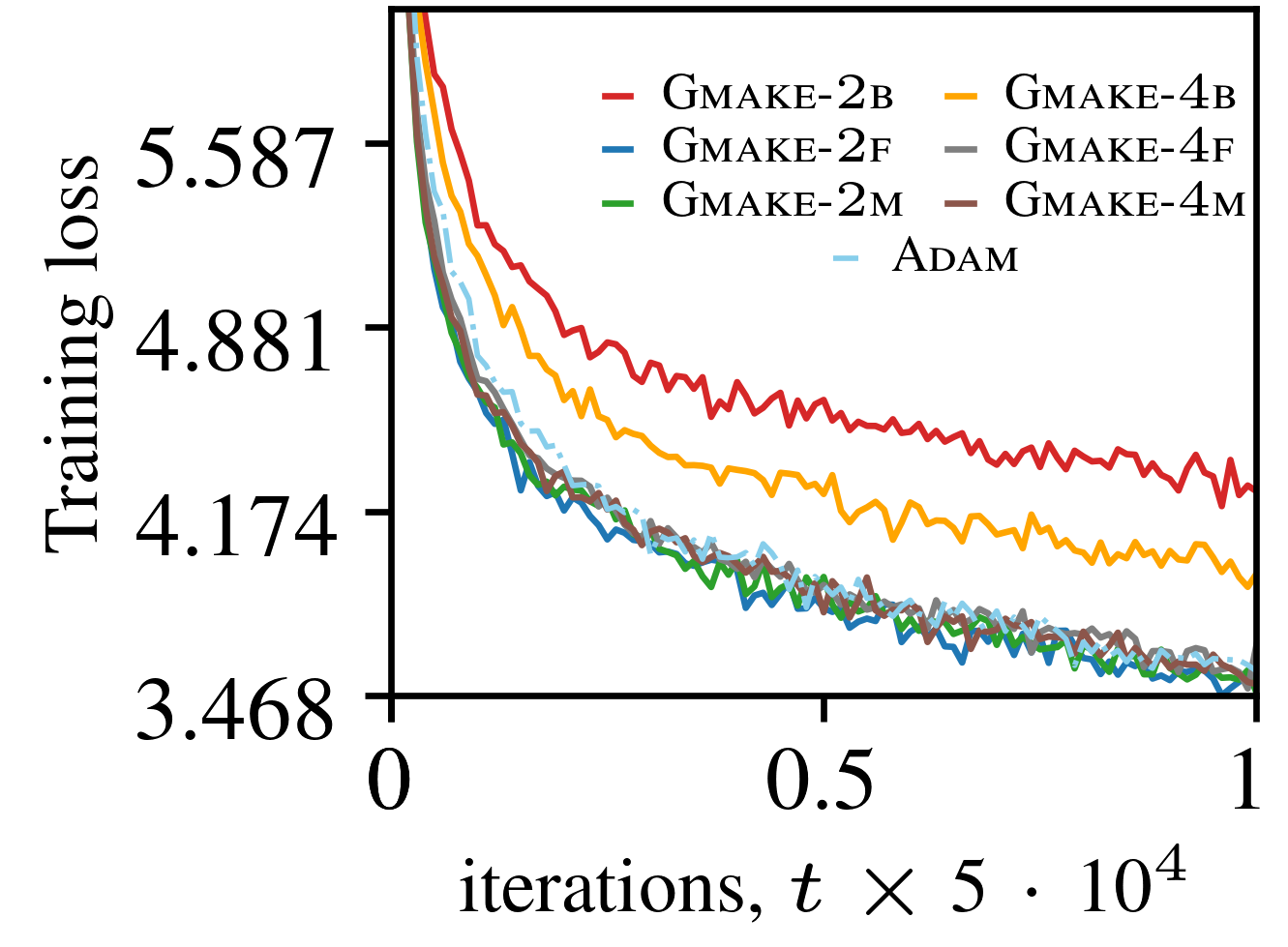}\hfil
  \includegraphics[width=0.48\textwidth]{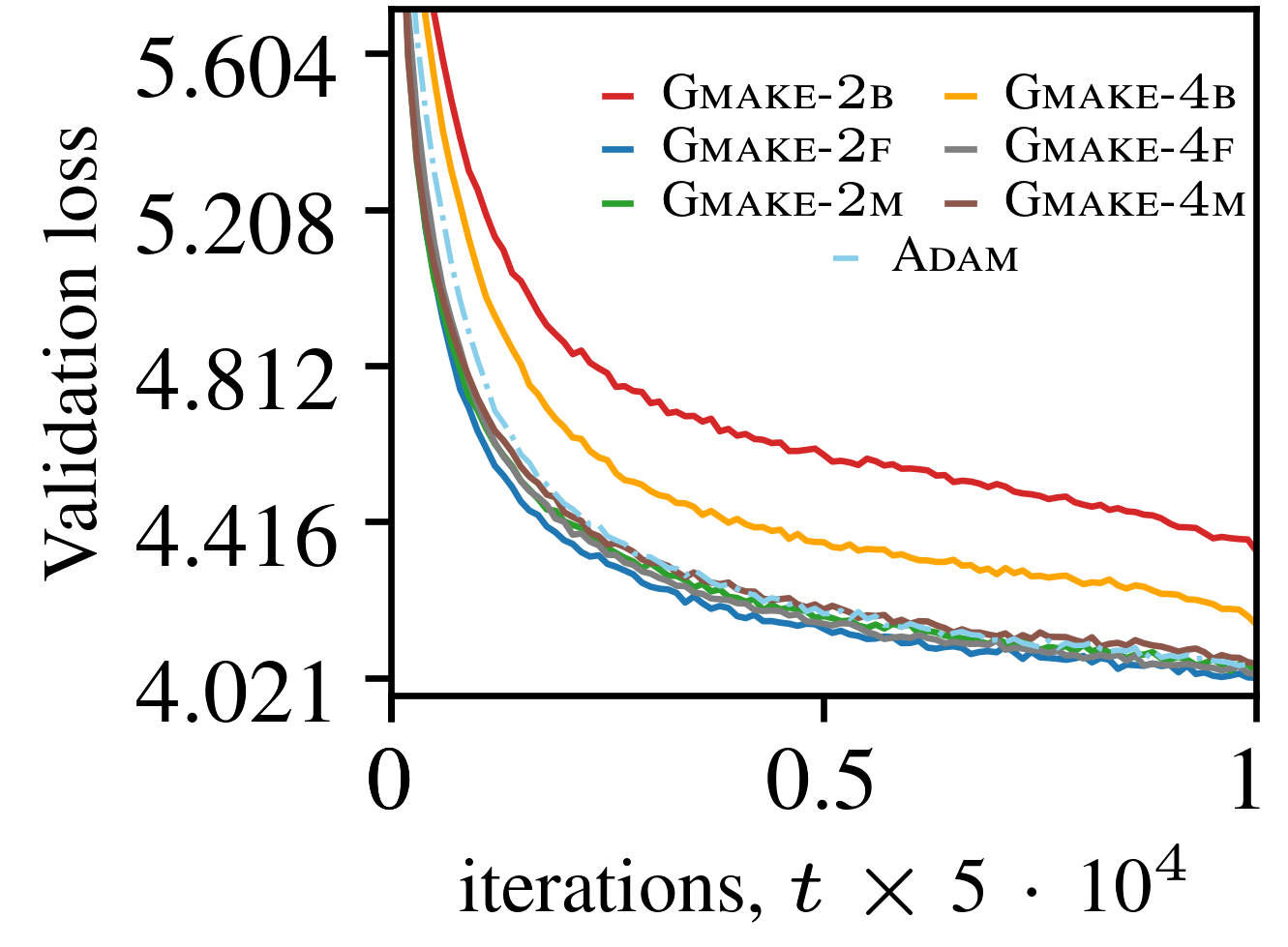}
  \caption{Training (left) and validation (right) loss for the basic (B),
spectrally filtered (F), and matrix-operator (M) forms of
\textsc{Gmake} with $p=2$ and $p=4$ on GPT2-124M trained on
FineWeb-Edu.}
  \label{fig:fweb50m}
\end{figure}
Figure~\ref{fig:fweb50m} compares the basic (B), filtered (F),
and matrix-operator (M) forms of \textsc{Gmake} for both
$p=2$ and $p=4$ on GPT2-124M trained on a 50 million token
subset of FineWeb-Edu. In all experiments, the same
maximum trust-region radius $\mu=5\times10^{-4}$, filter parameter
$\beta=0.9$, averaging coefficient $\rho=0.999$, and linear
decay schedule were used.

First, the filtered realizations substantially improve both
training and validation performance relative to their
corresponding basic forms. This is consistent with the
analysis of Section~3, where spectral lowpass regularization (momentum)
acts as a trust-region-preserving mechanism that improves the
quality of the gradient signal before update generation.

Second, within the basic realization, the fourth-moment form
consistently outperforms the corresponding second-moment form.
This suggests that, when a relatively large trust-region
radius is permitted, the additional higher-order moment
normalization provides beneficial update control beyond that
obtained from second-moment normalization alone.

Third, the matrix-operator realizations remain competitive
with their corresponding filtered forms despite satisfying a
stronger operator-level trust-region constraint. Since all
realizations use the same maximum trust-region radius
$\mu$, the matrix-operator form provides the strongest
trust-region guarantees among the three realizations while
maintaining comparable optimization performance.

\subsection{TinyStories}

\begin{figure}[h]
  \centering
  \includegraphics[width=0.48\textwidth]{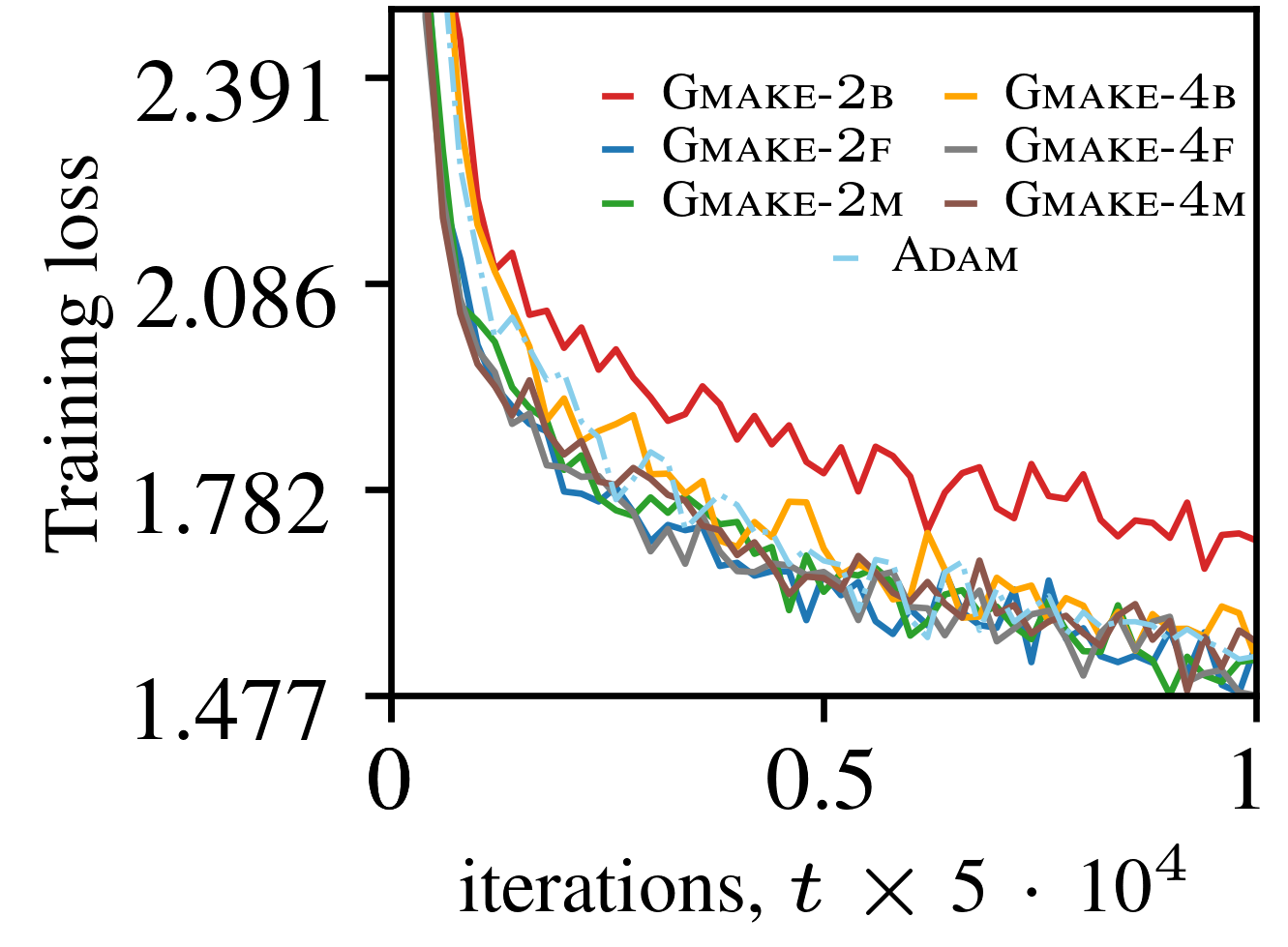}\hfil
  \includegraphics[width=0.48\textwidth]{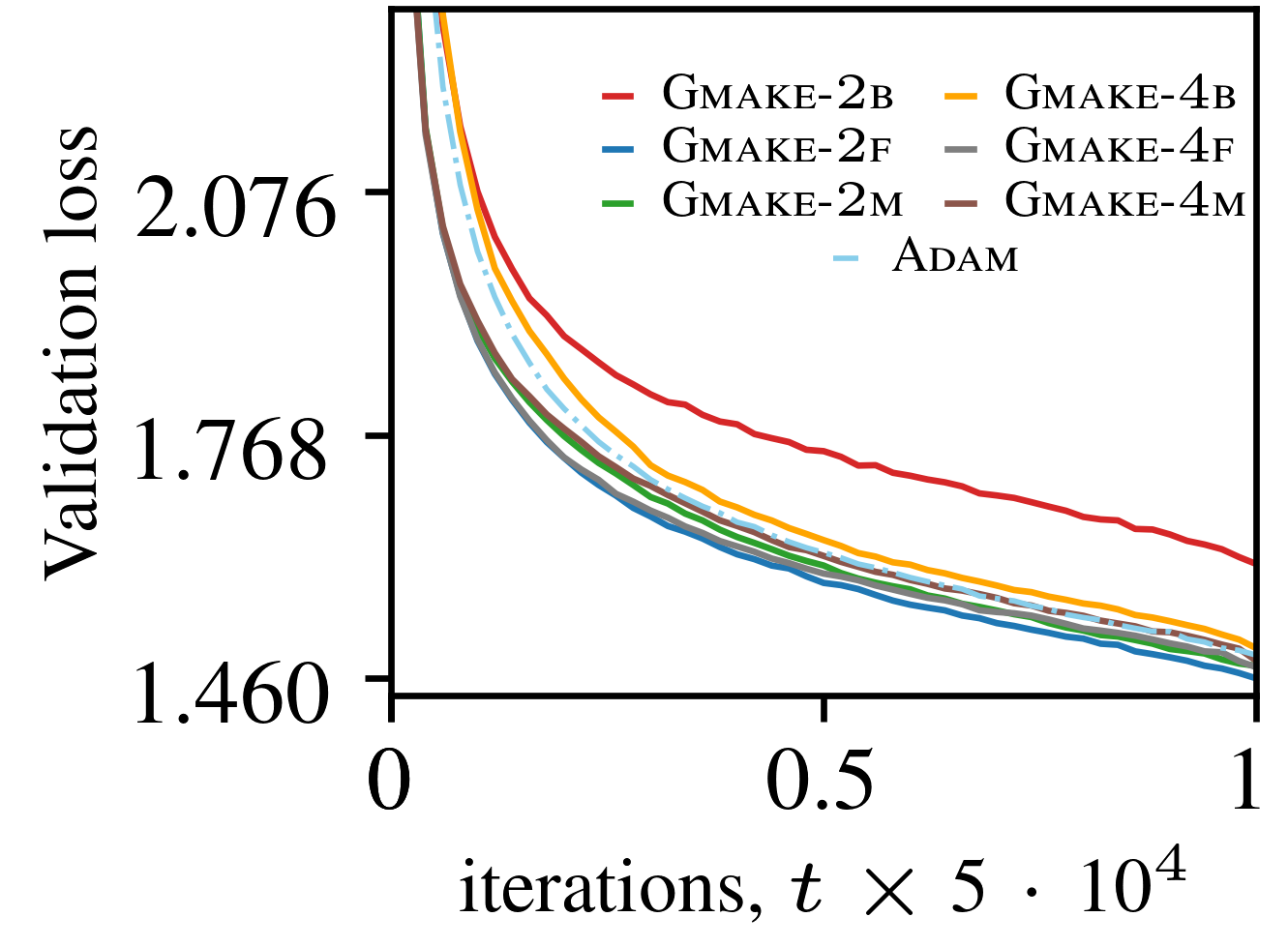}
  \caption{Training (left) and validation (right) loss for the basic (B),
spectrally filtered (F), and matrix-operator (M) forms of
\textsc{Gmake} with $p=2$ and $p=4$ on GPT2-124M trained on
TinyStories.}
  \label{fig:tstories}
\end{figure}
The results on TinyStories largely mirror those observed on
FineWeb-Edu. The filtered realizations again outperform their
corresponding basic forms, supporting the view that spectral
lowpass regularization provides a useful strengthening of the
underlying trust-region framework.

The fourth-moment realization again achieves lower losses
than the second-moment realization in the basic form.
However, after the introduction of filtering and
matrix-operator normalization, the performance gap between
the second- and fourth-moment realizations becomes
considerably smaller.

Across both datasets, the filtered and matrix-operator
training-loss trajectories are nearly indistinguishable,
while their corresponding validation losses remain similarly
close. These observations suggest that progressively stronger
trust-region controls can be imposed without materially
degrading optimization performance.

\subsection{Effect of Strengthening the Trust-Region}
\begin{figure}[h]
  \centering
  \includegraphics[width=0.48\textwidth]{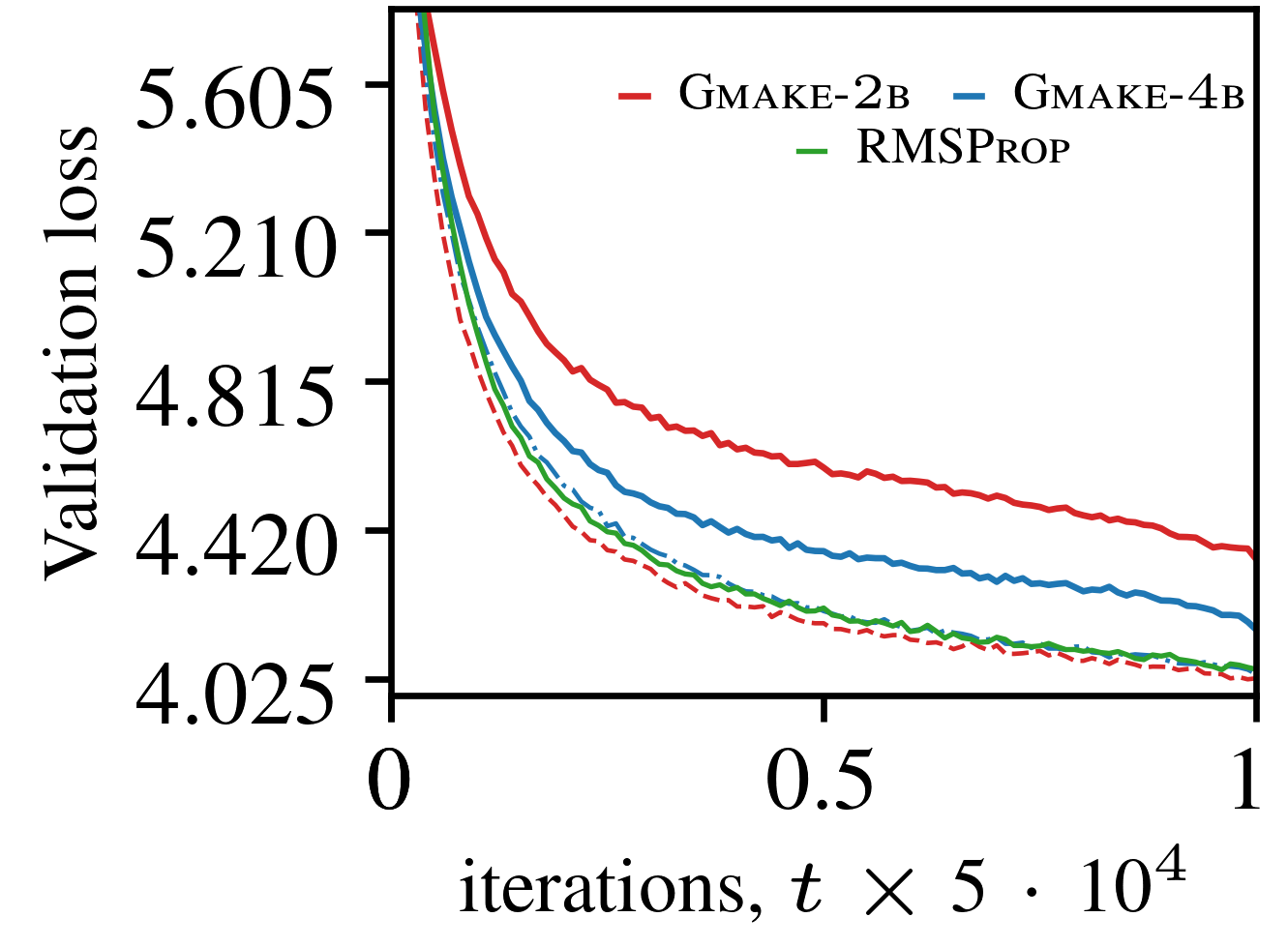}\hfil
  \includegraphics[width=0.48\textwidth]{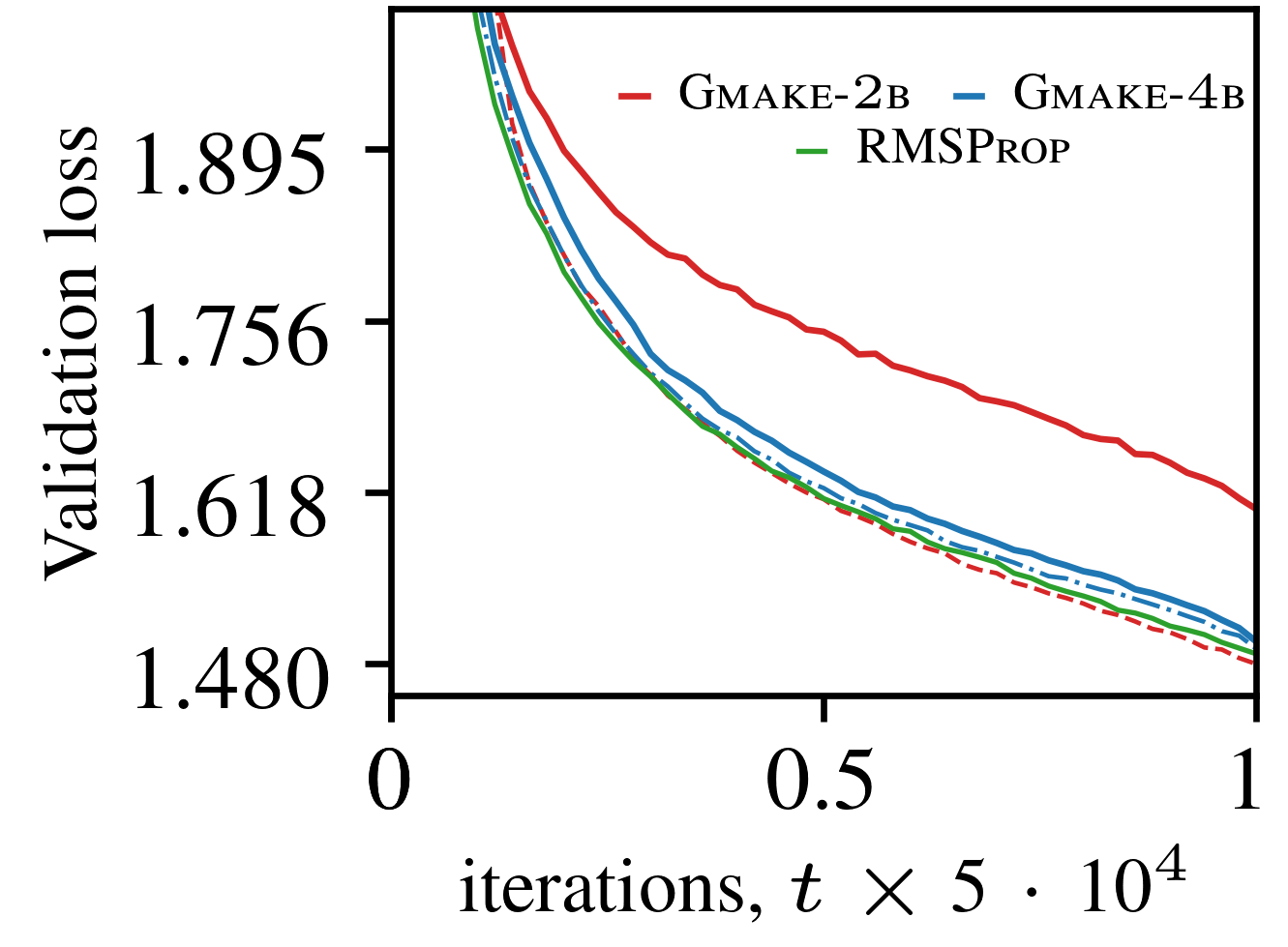}
  \caption{Validation loss on GPT2-124M trained on FineWeb-Edu (left) and
TinyStories (right). Solid lines correspond to the basic
\textsc{Gmake} realization with trust-region radius
$\mu=5\times10^{-4}$, while dash-dotted lines correspond to
$\mu=3\times10^{-4}$. \textsc{RMSProp} is included only for
$\mu=3\times10^{-4}$ because training with
$\mu=5\times10^{-4}$ was unstable. The larger trust-region
radius highlights the advantage of the fourth-moment
realization over the second-moment realization, while reducing
the trust-region radius substantially narrows the performance
gap between the two forms, with the second-moment realization becoming slightly better.}
  \label{fig:cmpts}
\end{figure}
Figure~\ref{fig:cmpts} investigates the effect of reducing the
maximum trust-region radius from
$\mu=5\times10^{-4}$ to $\mu=3\times10^{-4}$ in the
basic realization.

A notable observation is that the performance advantage of
the fourth-moment realization becomes substantially smaller
as the trust-region radius is reduced. For
$\mu=5\times10^{-4}$, the fourth-moment realization
consistently outperforms the second-moment realization on
both datasets. However, at the smaller trust-region radius,
the performance gap narrows considerably and
the second-moment realization achieves slightly lower
validation loss.

This behavior is consistent with the trust-region
interpretation developed in Section~2. The higher-order
normalization mechanism becomes most useful when larger
update magnitudes are permitted. As the trust-region radius
is reduced, update magnitudes are already more tightly
constrained, leaving less opportunity for the higher-order
moment normalization to provide additional benefit.

For reference, \textsc{RMSProp} was also evaluated at
$\mu=3\times10^{-4}$. Training with
$\mu=5\times10^{-4}$ was found to be substantially less
stable and is therefore omitted from the comparison. At the
smaller trust-region radius, the performance of \textsc{RMSProp} and
the second-moment realization become comparable.

\subsection{Trust-Region Insights}

The preceding experiments suggest several observations that are
best understood through the trust-region interpretation
developed in Sections~\ref{sec-gmake}-\ref{sec-spectral-mat} of this paper.

First, for the larger trust-region radius
$\mu=5\times10^{-4}$, the fourth-moment realization
consistently outperforms the corresponding second-moment
realization in the basic form on both FineWeb-Edu and
TinyStories. This behaviour is consistent with the
interpretation that the normalized fourth-moment estimator
provides additional update control beyond second-moment
normalization alone when relatively large update magnitudes are
permitted.

Second, the introduction of progressively stronger trust-region
controls appears to reduce the performance gap between the
second-moment and fourth-moment realizations. This effect is
observed after the introduction of spectral lowpass
regularization, matrix-operator normalization, and also when
the maximum trust-region radius is reduced from
$\mu=5\times10^{-4}$ to $\mu=3\times10^{-4}$ in the basic
realization.

Figure~\ref{fig:cmpts} provides additional evidence for this
observation. As the maximum trust-region radius is reduced, the
advantage of the fourth-moment realization becomes
substantially smaller, with the second-moment and fourth-moment
realizations achieving comparable performance on both
datasets. This suggests that the benefits of higher-order
moment normalization are most pronounced when the admissible
trust region is relatively large.

Taken together, these results indicate that the relative
advantage of the fourth-moment realization may be inversely
related to the strength of the surrounding trust-region
controls. When the trust-region constraints are weak, the
additional higher-order normalization provides a more
meaningful correction to the update process. As progressively
stronger trust-region controls are introduced through a
smaller trust-region radius, spectral filtering, or
operator-level normalization, the second-moment realization
appears sufficient to capture a larger fraction of the
achievable performance gains.

Finally, \textsc{RMSProp} was observed to train less stably than
its \textsc{Gmake} equivalent at the larger trust-region radius
$\mu=5\times10^{-4}$, and therefore is reported only for
$\mu=3\times10^{-4}$. This observation is consistent with the
trust-region interpretation developed in this paper, where the
additional normalization mechanisms of \textsc{Gmake} permit
stable optimization under larger admissible update magnitudes.

\section{Limitations, Gaps and Future Work}\label{sec-lim}

Although this trust-region framework establishes theoretical connections
between moment estimation, learning-rate scheduling,
spectral filtering, and matrix-operator normalization, several
questions remain open.

A notable empirical observation is that the performance advantage of the fourth-moment realization is most pronounced when the trust-region constraints are relatively weak. In the basic realization, operated with a larger trust-region radius, the fourth-moment form initially
outperformed its second-moment counterpart. However, this advantage diminished substantially as stronger trust-region controls were introduced, whether through spectral filtering, matrix-operator normalization, or a smaller maximum trust-region radius. This trend suggests that part of the benefit provided by fourth-moment normalization may overlap with the stabilization and regularization effects already supplied by these stronger trust-region mechanisms.

The precise cause of this behavior remains unclear. One
possible explanation is that these mechanisms partially
address the same update fluctuations that the higher-order
moment normalization is designed to regulate, thereby
reducing its incremental benefit. Alternatively, the
observation may indicate that spectral lowpass
regularization and spectral trust-region control naturally
strengthen second-moment mechanisms more than higher-order
moment mechanisms.

Nonetheless, this observation may expose a gap in the current framework. While spectral lowpass regularization appears to provide a natural strengthening of the second-moment trust-region mechanism, a corresponding regularization principle specifically aimed at higher-order moment control has not yet been identified. It therefore remains unclear whether an analogous mechanism exists and, if so, whether it could recover or further enhance the benefits of the $p=4$ realization. More broadly, a systematic investigation of the performance and stability characteristics across the range $1 < p \le 4$, especially the regime $p < 2$, may provide deeper insight into the role of moment order in trust-region control and accelerated learning.

Moreover, the numerical experiments were intended primarily as proof-of-concept demonstrations rather than a comprehensive empirical evaluation. For consistency across experiments, all realizations were evaluated using the same values of $\mu$, $\beta$, and $\rho$, and only two representative choices of $p$ were examined. 
An important direction for future work is therefore a systematic investigation of the sensitivity of the basic, filtered, and matrix-operator formulations to these hyperparameters. Such studies could also explore alternative normalization orderings, adaptive selections of $p$, and spectral-level trust-region constructions across a wider range of learning problems.


\section{Conclusion}\label{sec:concl}
This paper developed a trust-region framework for studying the moment
estimation mechanism in stochastic gradient optimization. The
derived family of learning-rate mechanisms generalizes
second-moment methods through $p$-th moment trust-region
constraints. Within this framework,
\textsc{RMSProp} and \textsc{Adam} can be interpreted respectively, as similar to the second-moment realizations of the basic and spectrally
filtered \textsc{Gmake} formulations.

The \textsc{Gmake} framework further reveals that
learning-rate scheduling, momentum, and spectral-norm
normalization can be interpreted as complementary
trust-region mechanisms acting on different properties of the
update process. By extending the basic realization through
spectral lowpass filtering and matrix-operator normalization,
progressively stronger trust-region controls are obtained
within a common framework. The experimental
results further suggest the existence of a trust-region
hierarchy, in which the incremental benefit of higher-order
moment normalization decreases as these progressively stronger
trust-region controls are imposed on the update process.

Taken together, these developments provide a unified
trust-region interpretation of several mechanisms commonly
used in optimizing deep neural networks via the stochastic
gradient algorithm.

\bibliography{refs}
\bibliographystyle{tmlr}

\appendix

\end{document}